\documentclass{article}

\usepackage[preprint]{neurips_2026}

\usepackage[utf8]{inputenc} % allow utf-8 input
\usepackage[T1]{fontenc}    % use 8-bit T1 fonts
\usepackage{hyperref}       % hyperlinks
\usepackage{url}            % simple URL typesetting
\usepackage{booktabs}       % professional-quality tables
\usepackage{amsfonts}       % blackboard math symbols
\usepackage{nicefrac}       % compact symbols for 1/2, etc.
\usepackage{microtype}      % microtypography
\usepackage{xcolor}         % colors
\usepackage{adjustbox}
\usepackage{amsmath}
\usepackage{multirow}
\usepackage{algorithm}
\usepackage{algpseudocode}
\usepackage{subcaption}
\usepackage{tablefootnote}

\title{Implicit Regularization of Mini-Batch Training in Graph Neural Networks}

\author{%
  Clement Wang
    \\
  Institut Polytechnique de Paris\\
  Paris, France \\
  \texttt{clement.wang@ip-paris.fr} \\
  \And
  Antoine Vialle \\
  Institut Polytechnique de Paris \\
  Paris, France \\
  \texttt{antoine.vialle@ip-paris.fr} \\
  \AND
  Robin Vaysse \\
  Mirakl \\
  Paris, France \\
  \texttt{robin.vaysse@mirakl.com} \\
  \And
  Thomas Bonald \\
  Institut Polytechnique de Paris \\
  Paris, France \\
  \texttt{thomas.bonald@ip-paris.fr} \\
}

\begin{document}

\maketitle

\begin{abstract}
    Mini-batch training of Graph Neural Networks (GNNs) is fundamentally different from training on i.i.d. data: sampling a subgraph alters the topology and introduces boundary effects, leading prior work to develop structure-aware samplers that preserve local connectivity and reduce embedding variance. Surprisingly, we demonstrate that the simplest possible scheme, Random Node Sampling (RNS) — training on the induced subgraph of uniformly sampled nodes — matches or outperforms full-graph training on 8 of 10 datasets at a fraction of the wall-clock time and memory. To explain this, we apply backward error analysis to graph mini-batch Stochastic Gradient Descent (SGD) and show that it implicitly minimizes the sampled loss plus a regularizer proportional to the mini-batch gradient variance, a quantity directly shaped by the sampler. Although RNS discards local structure, it produces mini-batches whose expected loss is closer to the full-graph loss, and whose per-batch gradients have lower variance, yielding a better implicit objective. Our analysis reframes the choice of graph sampler as a form of implicit regularization, and identifies RNS as a strong, theoretically grounded method for scalable GNN training.
\end{abstract}

\section{Introduction}

Graph Neural Networks (GNNs) are commonly trained transductively, with the full graph available during optimization \cite{kipf2016semi, velivckovic2017graph}. While conceptually simple, full-graph training becomes costly at scale: message passing couples many node representations, increasing memory and compute requirements and limiting the number of optimization steps possible under a fixed budget. This has motivated sampling-based and mini-batch training methods for GNNs \cite{hamilton2017inductive, zeng2019graphsaint}.

Graph mini-batching, however, is fundamentally different from mini-batching i.i.d. examples. Since node predictions depend on neighboring features and edges, training on a sampled subgraph changes the interactions available to the model and introduces boundary effects. The sampler is therefore not merely an implementation detail: it changes the stochastic gradients, and hence the effective learning problem. This raises a basic question: when a GNN is trained on sampled subgraphs, what objective is  actually optimized?

This paper studies \emph{Random Node Sampling} (RNS) for transductive node classification. At each iteration, RNS samples nodes uniformly at random and trains on the induced subgraph. RNS has recently appeared in graph transformers \cite{wu2022nodeformer, wu2023sgformer, Deng2024PolynormerPG} and message-passing GNNs \cite{luo2024classic}, but mainly as a practical mechanism for reducing memory and runtime. As a result, its effect on predictive performance has remained entangled with architectural and optimization choices.

Beyond computational savings, graph mini-batching introduces a form of implicit regularization that has received comparatively little attention. Prior work has focused on the subgraph approximation effect — whether sampled subgraphs faithfully recover the full-graph objective — while the optimization effect induced by stochastic updates on structured batches has largely been overlooked. Using backward error analysis, we disentangle these two effects under SGD and show that training on sampled subgraphs implicitly minimizes a modified objective: the sampled loss augmented by a regularizer proportional to the variance of mini-batch gradients. We then show that RNS is distinctive in this regard — despite constructing batches without any knowledge of graph structure, it produces per-batch gradients whose statistics closely match those of the full graph, yielding a lower-variance, better-controlled implicit objective than structure-aware alternatives.

\paragraph{Contributions.}
\begin{itemize}
    \item \textbf{Theory.} We apply backward error analysis to GNN mini-batch training and
    prove that SGD over sampled subgraphs implicitly minimizes a modified objective combining
    the sampled loss with a gradient-variance regularizer shaped by the sampler. We show that
    RNS induces a well-controlled bias relative to $L_{\mathrm{full}}$ and a gradient-variance
    regularization term orders of magnitude smaller than that of structure-based samplers,
    yielding a more stable implicit objective.

    \item \textbf{Benchmark.} We benchmark five
    sampling strategies across three large-scale graphs and four GNN architectures. RNS
    matches or outperforms full-graph training on 8 of 10 benchmarks spanning diverse graph
    sizes, densities, and homophily levels.

    \item \textbf{Practical impact.} RNS is a drop-in replacement for full-graph training
    requiring only a single integer hyperparameter $m$. It achieves 2$\times$ to
    12$\times$ wall-clock speedups and up to 3$\times$ lower peak GPU memory,
    making it a strong default for scalable transductive GNN training.
\end{itemize}

\section{Related work}

\paragraph{GNNs and large-scale training.}
Scaling GNNs to large graphs relies on mini-batch training methods. Neighborhood sampling recursively samples local neighborhoods \cite{hamilton2017inductive, ying2018graph}, while layer-wise methods such as FastGCN \cite{chen2018fastgcn}, LADIES \cite{zou2019layer}, and LABOR \cite{balin2023layer} reduce cost by sampling nodes at each layer. Subgraph-based methods such as ClusterGCN \cite{chiang2019cluster} and GraphSAINT \cite{zeng2019graphsaint} process graph partitions or induced subgraphs as mini-batches. Random node sampling has recently appeared as a practical mechanism in graph transformers \cite{wu2022nodeformer, wu2023sgformer, Deng2024PolynormerPG} and message-passing GNNs \cite{luo2024classic}, but its effect on optimization dynamics has not been systematically analyzed. These methods are primarily motivated by efficiency; how the resulting mini-batch noise shapes the effective training objective remains an open question.

\paragraph{Optimization dynamics and positioning.}
A complementary line of work studies stochastic optimization in deep learning via continuous-time approximations, modeling SGD using stochastic differential equations \cite{Mandt2017SDE, Smith2018SDEBay, Jastrzebski2017SDEthree, chaudhari2018stochastic} or backward analysis \cite{Hairer2006, Barrett2021implicitGD, Cattaneo2024implicitADAM, ghosh2023implicit}. \citet{Smith2021implicitSGD} applies backward analysis to SGD and shows that its iterates track the gradient flow of a modified objective, which can be interpreted as the original loss augmented with an implicit regularizer.
These frameworks clarify how discretization and stochasticity induce implicit regularization, but they are typically developed under i.i.d.\ data assumptions. In contrast, we analyze the \emph{structured} stochasticity induced by graph mini-batch training, and we characterize how Random Node Sampling modifies the effective training objective of GNNs.

\begin{figure*}[t]
\centering

\resizebox{\textwidth}{!}{%
\begin{tabular}{c c c c c}
 & \begin{tabular}[c]{@{}c@{}}
     \textbf{Neighborhood Sampling} \\
     \scriptsize fanout = [2], batch size = 3
   \end{tabular}
 & \begin{tabular}[c]{@{}c@{}}
     \textbf{ClusterGCN} \\
     \scriptsize num clusters = 4, num clusters per batch = 2
   \end{tabular}
 & \begin{tabular}[c]{@{}c@{}}
     \textbf{GraphSAINT} \\
     \scriptsize walk length = 7, num seeds = 1
   \end{tabular}
 & \begin{tabular}[c]{@{}c@{}}
     \textbf{Random Node Sampling} \\
     \scriptsize num parts = 2
   \end{tabular}
 \\[0.8em]

\textbf{Sampling process}
 & \adjustbox{valign=c}{\includegraphics[height=3.2cm]{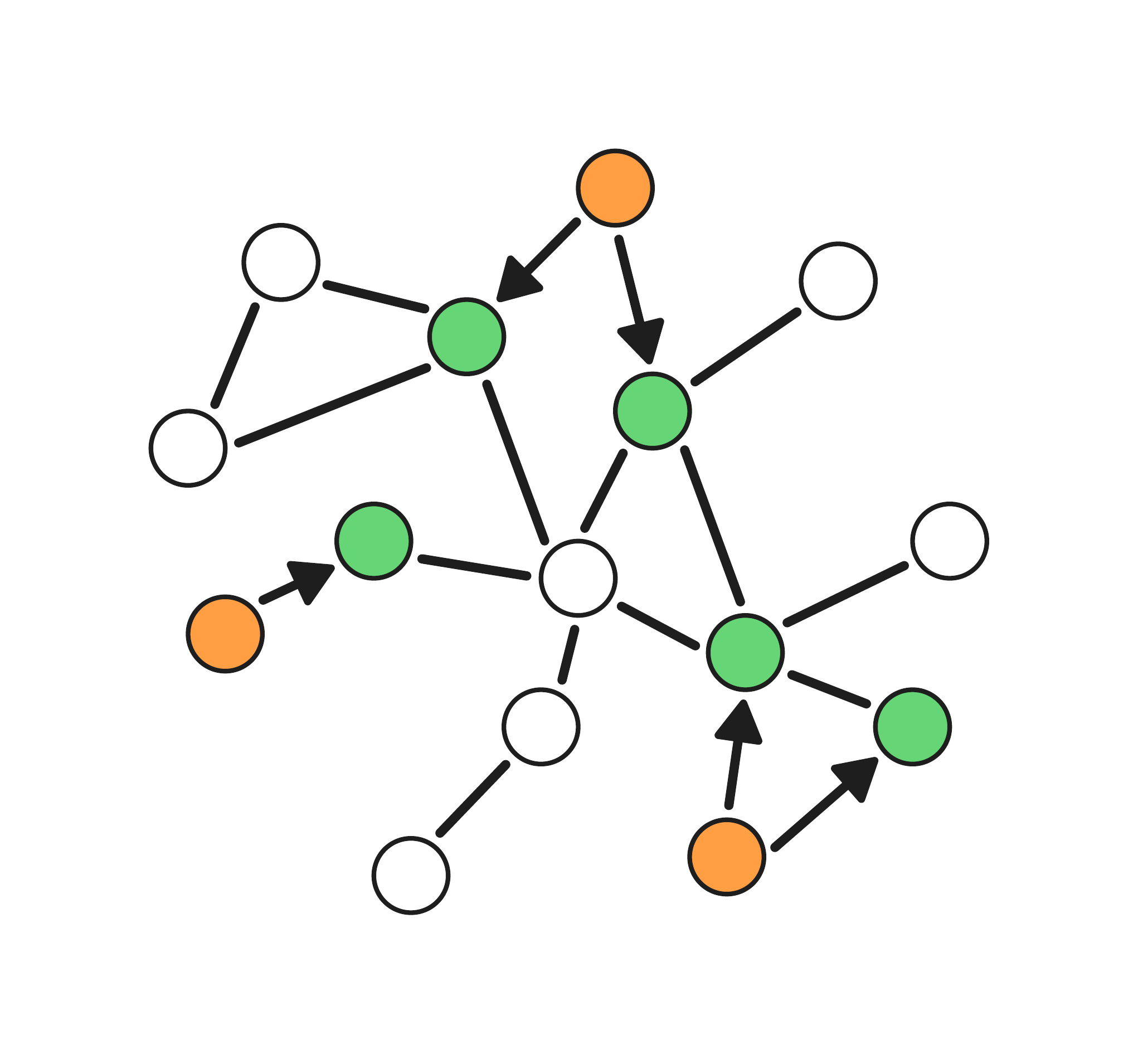}}
 & \adjustbox{valign=c}{\includegraphics[height=3.2cm]{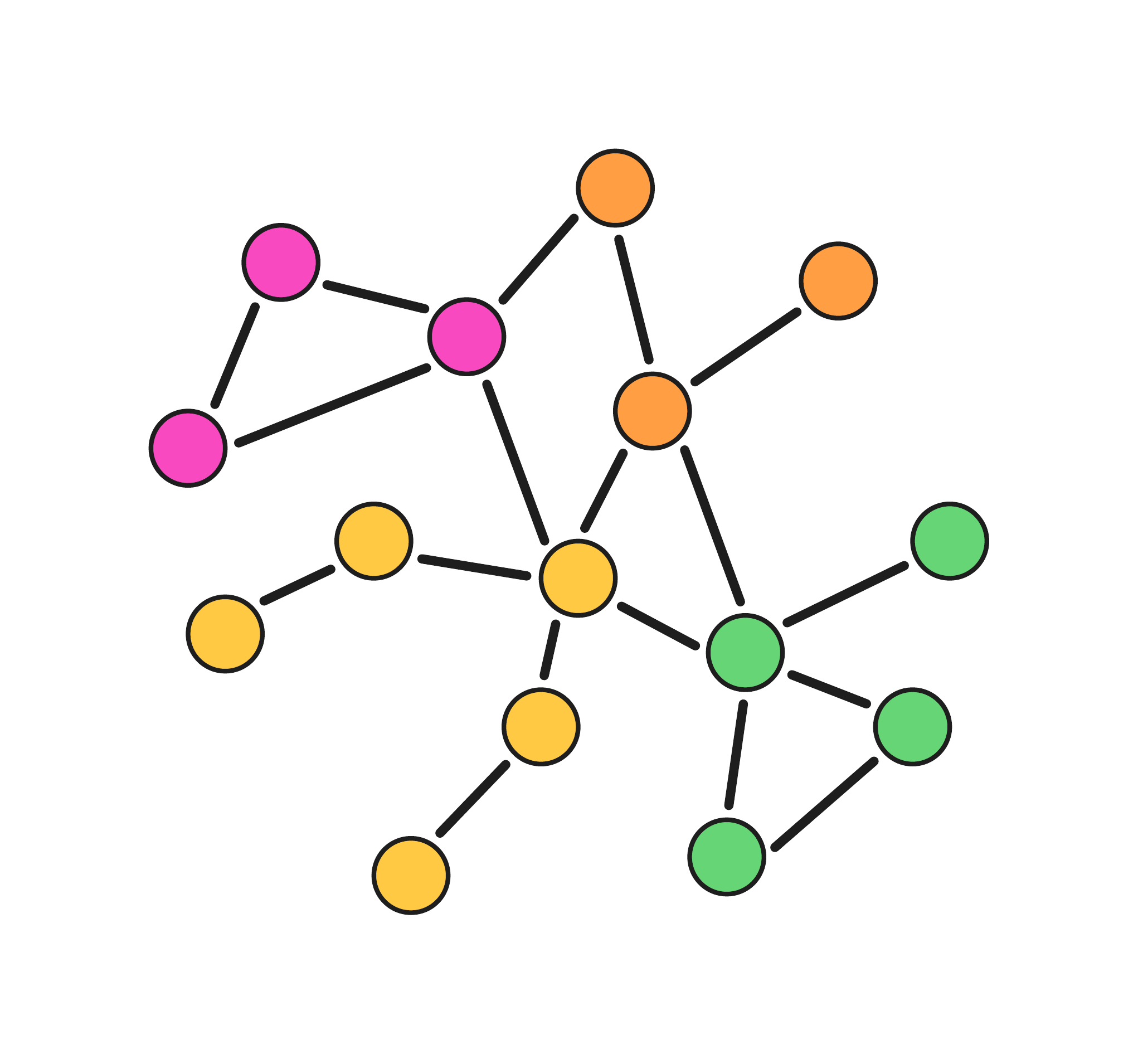}}
 & \adjustbox{valign=c}{\includegraphics[height=3.2cm]{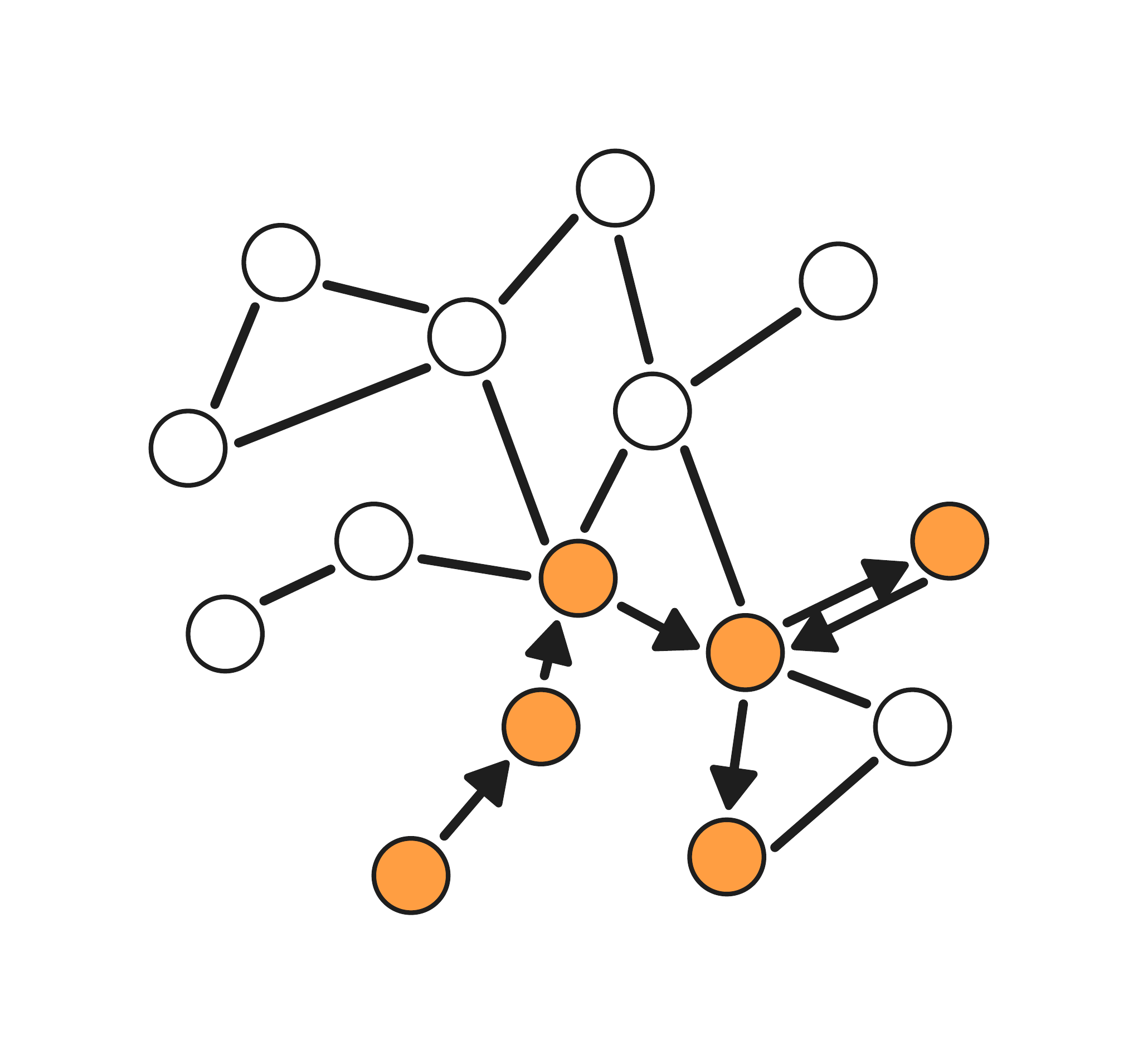}}
 & \adjustbox{valign=c}{\includegraphics[height=3.2cm]{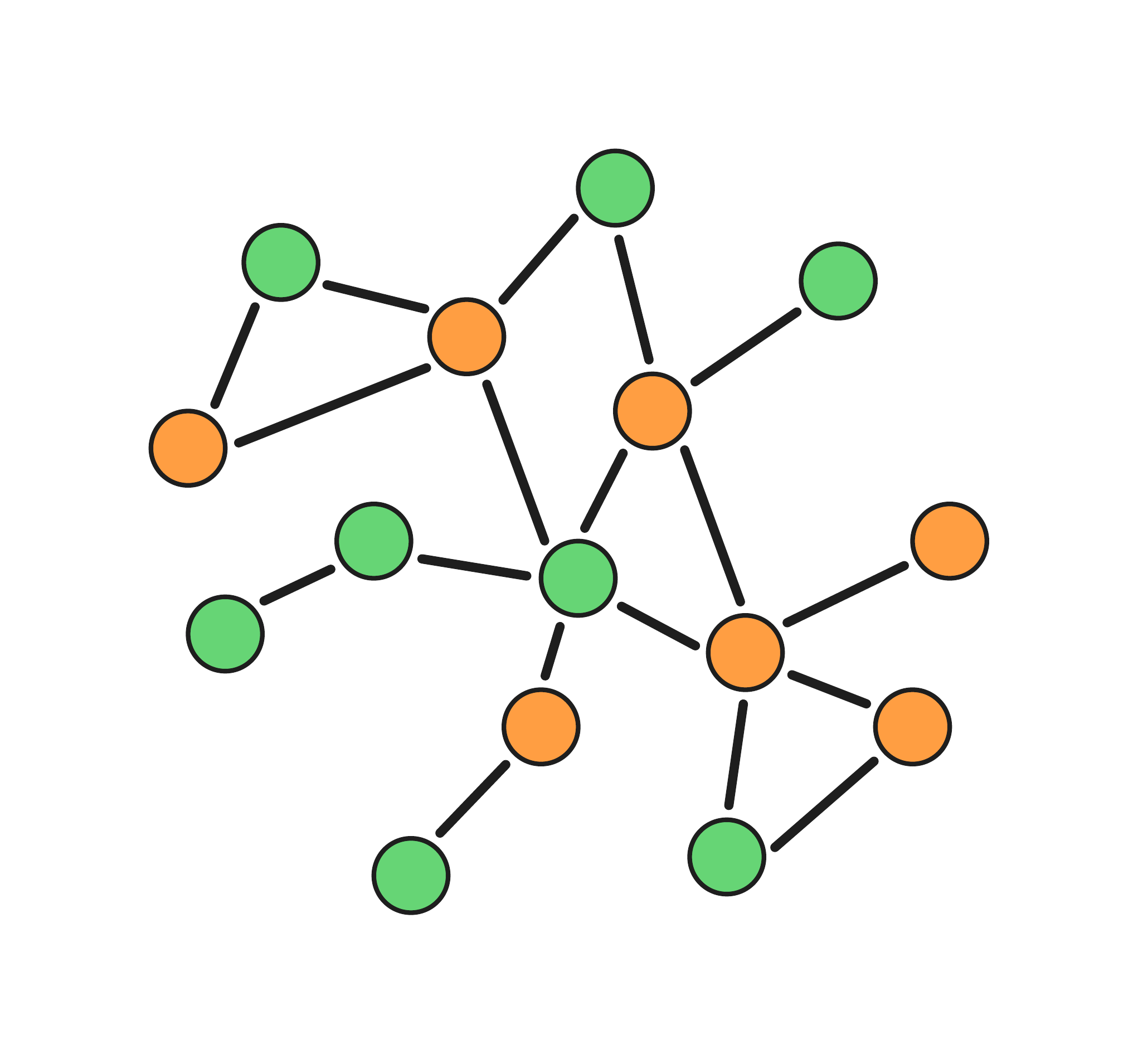}} \\[0.9em]

\textbf{Sampled subgraph}
 & \adjustbox{valign=c}{\includegraphics[height=3.2cm]{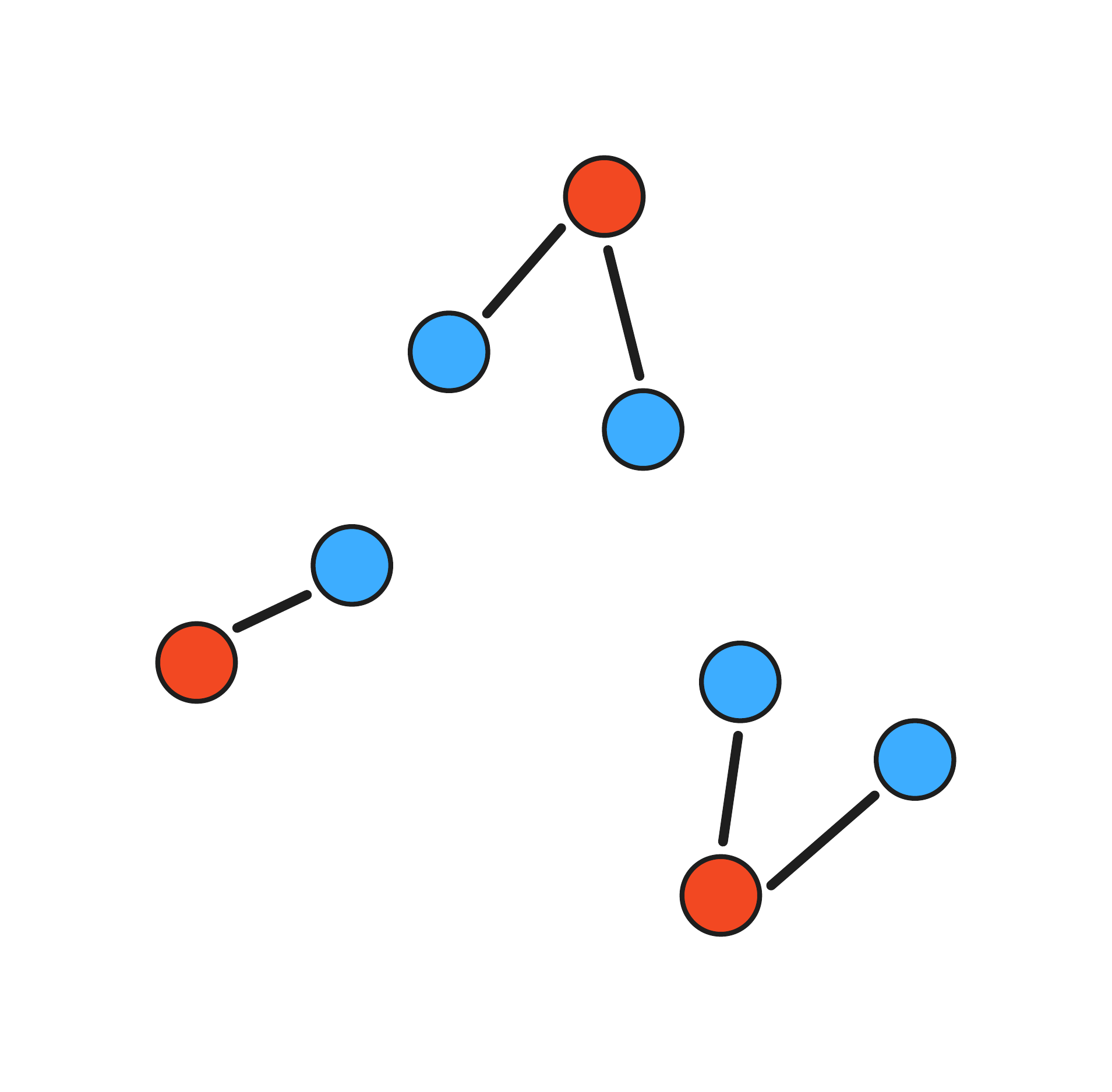}}
 & \adjustbox{valign=c}{\includegraphics[height=3.2cm]{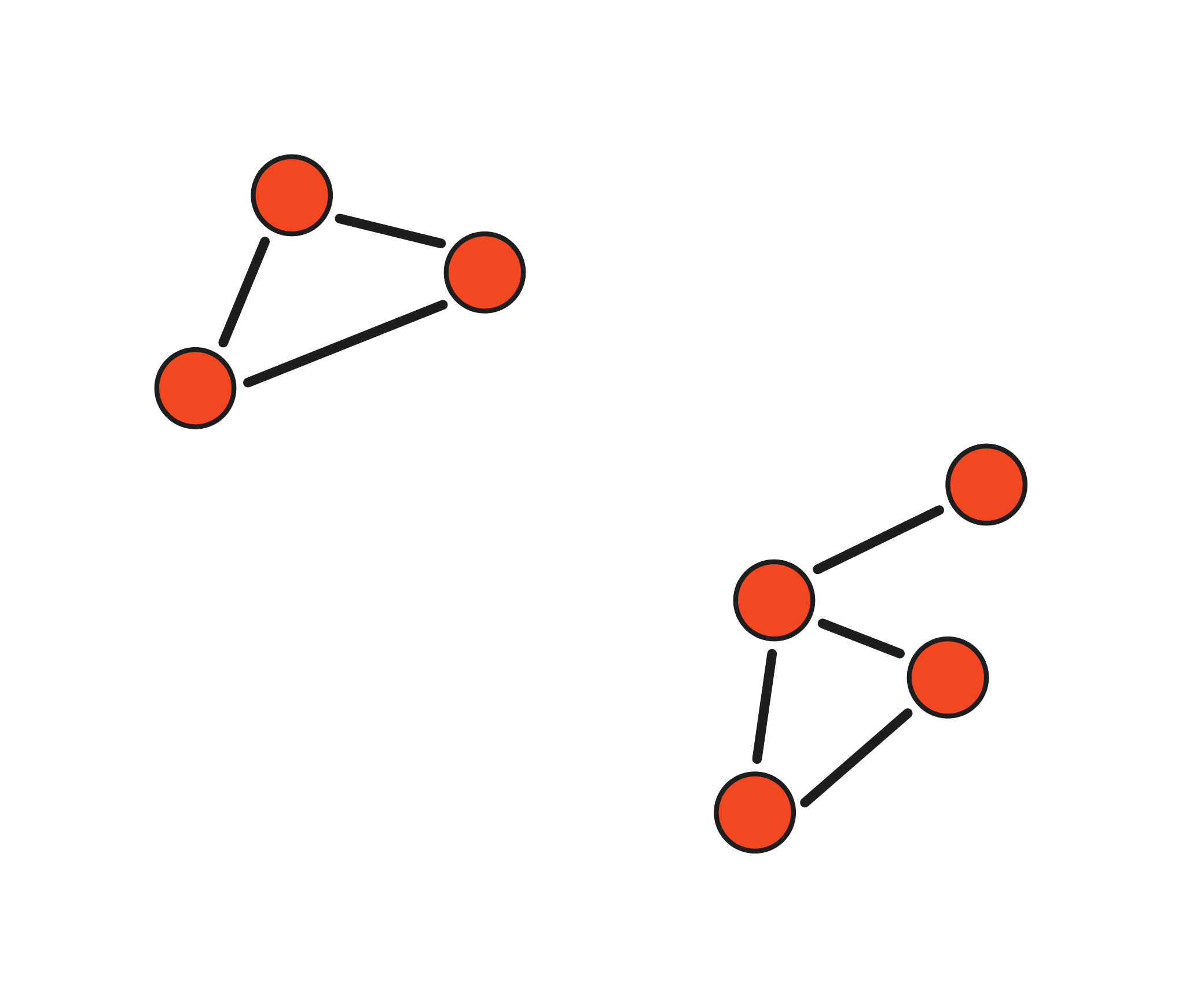}}
 & \adjustbox{valign=c}{\includegraphics[height=3.2cm]{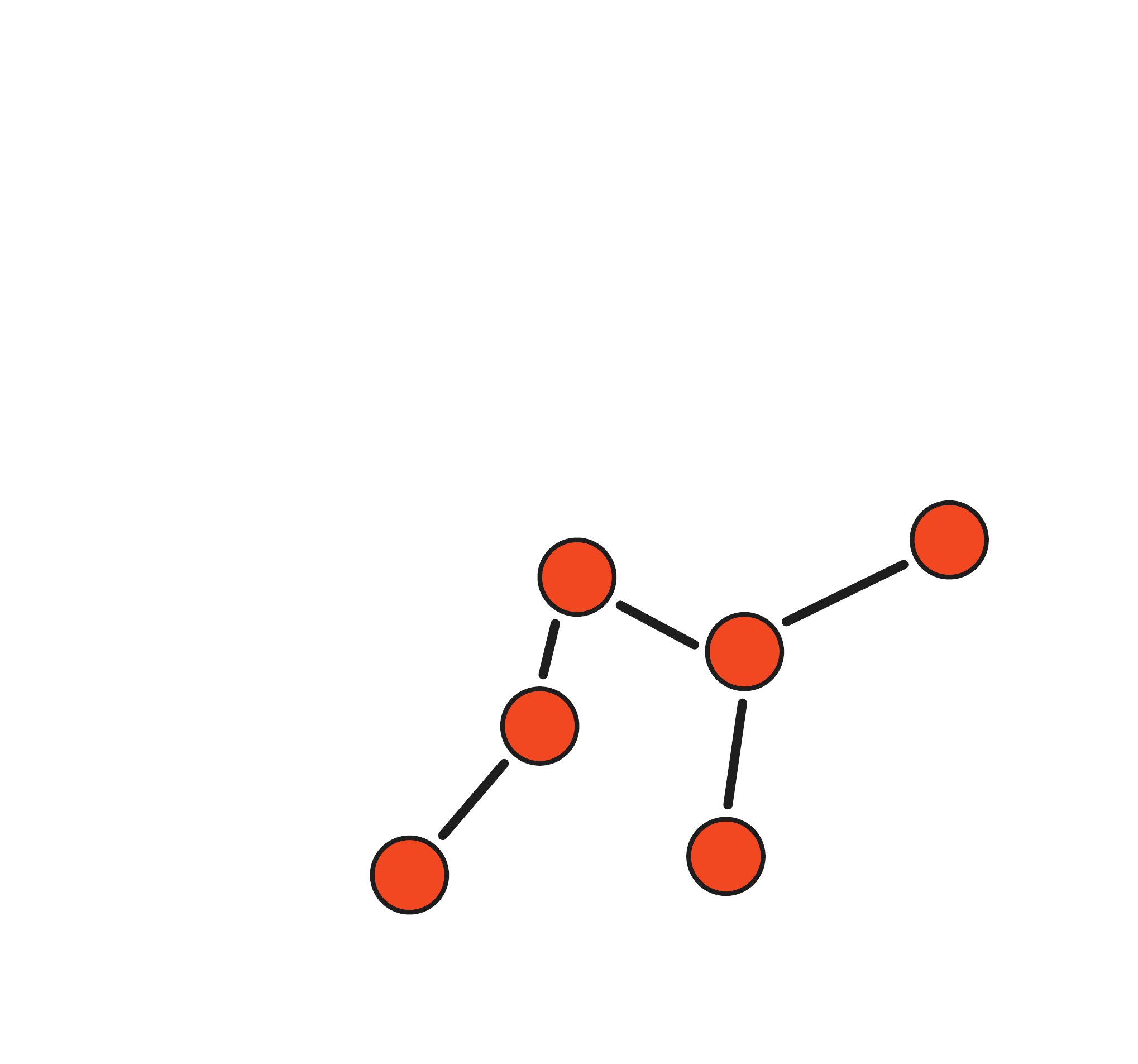}}
 & \adjustbox{valign=c}{\includegraphics[height=3.2cm]{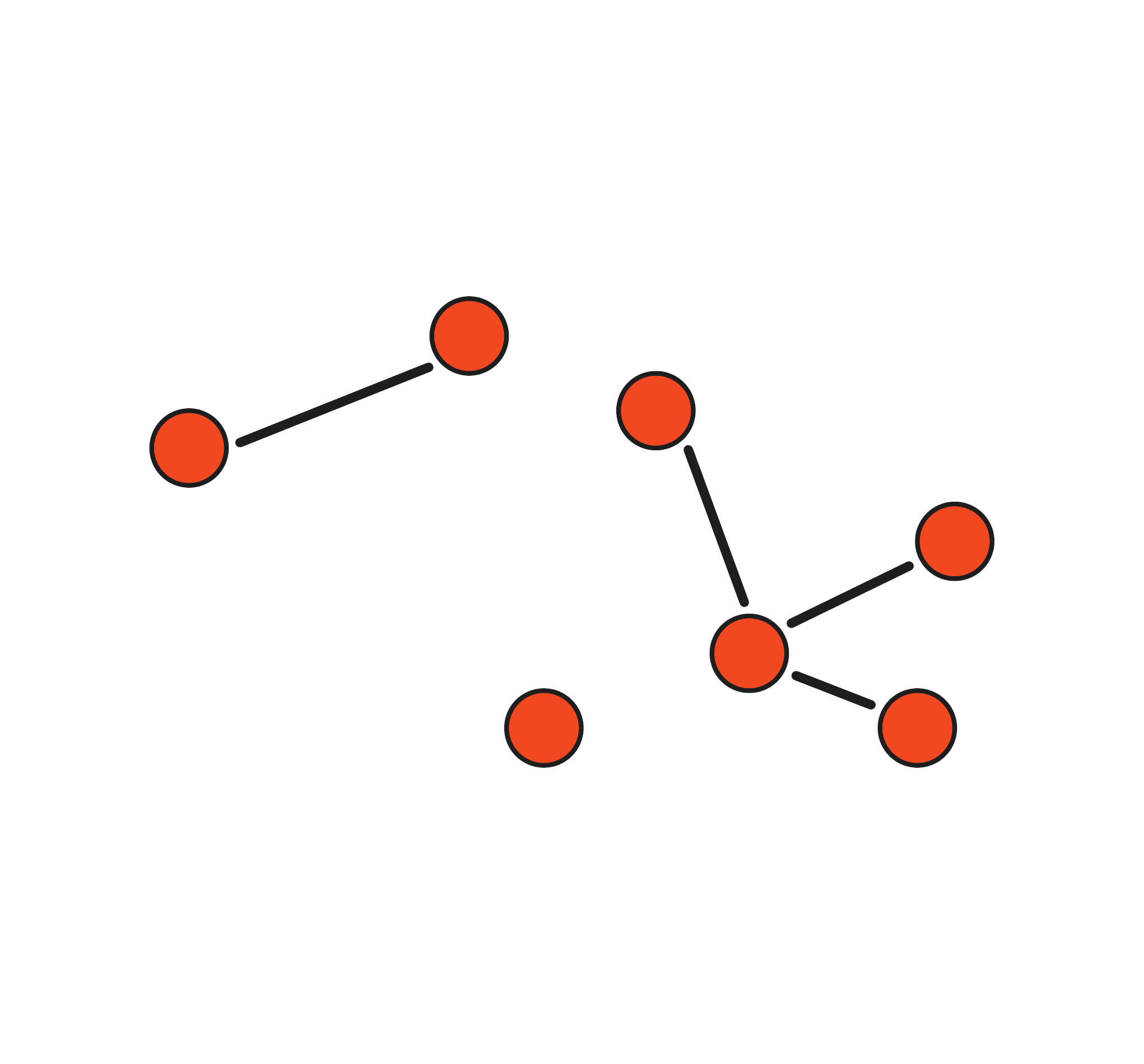}}
\end{tabular}%
}

\caption{%
\textbf{Batch sampling strategies for GNN training.}
Top: mini-batch construction for each method. Bottom: resulting sampled subgraph.
\textcolor{red}{Red} nodes are targets ($B^{\mathrm{train}}$), while \textcolor{blue}{blue} nodes are auxiliary nodes used only for message passing ($B \setminus B^{\mathrm{train}}$). See Appendix~\ref{subsec:sampling_strategies} for details.
}
\label{fig:gnn_sampling}
\end{figure*}

\section{Setting, notation, and sampling strategies}
We study \emph{transductive} node-level learning on a graph $G=(V,E)$, where each node $v\in V$ has features $x_v\in\mathbb{R}^d$ and a label $y_v$. The node set is split into disjoint subsets
\[
V = V^{\mathrm{train}} \,\dot\cup\, V^{\mathrm{val}} \,\dot\cup\, V^{\mathrm{test}}.
\]
In the transductive regime, the learner has access to the full graph structure $E$ and all node features $X=\{x_v\}_{v\in V}$ during training, but the supervised loss is evaluated only on training nodes.

Let $f(w, ., .)$ denote a graph neural network (GNN) with parameters $w$. For a node $v$, we write $f(w, X,E)[v]$ the model output at node $v$ when evaluated on the full graph. Using a generic per-node loss $\ell(\cdot,\cdot)$, \emph{full-graph training} minimizes
\begin{equation}
\label{eq:full_loss}
L_{\mathrm{full}}(w)
\;=\;
\frac{1}{|V^{\mathrm{train}}|}
\sum_{v\in V^{\mathrm{train}}}
\ell\!\big(f(w, X,E)[v],\,y_v\big).
\end{equation}

\paragraph{Mini-batch subgraph training.}
To scale to large graphs, one often replaces the full-graph forward pass by a sampled subgraph. A mini-batch is a subgraph
\[
G_{B}=(B,E_{B}),
\qquad
X_{B}=\{x_v\}_{v\in B},
\]
together with a set of \emph{supervised target} nodes $B^{\mathrm{train}}\subseteq B\cap V^{\mathrm{train}}$ on which the loss is computed. The corresponding mini-batch objective is
\begin{equation}
\hat L_{\mathrm{batch}}^{B}(w)
\;=\;
\frac{1}{|B^{\mathrm{train}}|}
\sum_{v\in B^{\mathrm{train}}}
\ell\!\big(f(w, X_B,E_B)[v],\,y_v\big).
\end{equation}

\paragraph{Random node sampling (RNS).}
In this work we focus on \emph{Random Node Sampling} (RNS). At the start of each epoch, we randomly split the node set $V$ into $m$ disjoint subsets $B_0,\dots,B_{m-1}$ of (approximately) equal size. Each training step then processes one subset $B_k$: we form the mini-batch graph as the induced subgraph $G[B_k]$, and compute the supervised loss only on the labeled training nodes within that batch, i.e., $B_k^{\mathrm{train}} = B_k \cap V^{\mathrm{train}}$ (Alg.~\ref{alg:rns}).

\paragraph{Other sampling strategies.}
Common alternatives include Neighborhood Sampling \cite{hamilton2017inductive}, ClusterGCN \cite{chiang2019cluster}, and GraphSAINT \cite{zeng2019graphsaint} (see Figure~\ref{fig:gnn_sampling} and Appendix~\ref{subsec:sampling_strategies}). These approaches construct mini-batches using local expansions, graph partitions, or subgraph sampling. We refer to the original papers for details.

Unlike other samplers, RNS does not explicitly preserve local structure: nodes typically have incomplete neighborhoods compared to $G$. One might therefore expect its stochastic gradients to differ substantially from those induced by structure-based mini-batches. We show, however, that despite this local sparsification, RNS yields mini-batch losses that closely approximate the full-graph loss, with low bias and variance.

\section{On the implicit learning objective of GNN training with backward analysis}

\subsection{Backward analysis of GNN training}
\label{sec:theory}

Backward analysis interprets a discrete optimization algorithm not as an
approximation of a continuous-time gradient flow, but as the \emph{exact}
discretization of a \emph{modified} flow. The modified system captures the
cumulative effect of discretization errors and reveals implicit biases induced
by the algorithm.

\paragraph{Gradient descent as a discretized ODE.}
Gradient descent with step size $\epsilon>0$ on a loss $L(w)$,
\begin{equation}
w_{k+1} = w_k - \epsilon \nabla L(w_k),
\end{equation}
is the forward Euler discretization of the gradient flow
$\dot w(t) = - \nabla L(w(t))$. For finite $\epsilon$, discretization errors
accumulate and the discrete trajectory no longer follows the exact flow.
Backward analysis seeks a modified vector field $\tilde f$ such that the iterates
$\{w_k\}$ lie exactly on the flow of $\dot w = \tilde f(w)$, up to a given order
in $\epsilon$. Assuming an asymptotic expansion
$\tilde f(w) = f(w) + \epsilon\,g(w) + O(\epsilon^2)$ with $f(w)=-\nabla L(w)$,
the correction $g$ encodes an \emph{implicit regularization} effect induced by
the finite step size~\citep{Hairer2006,Smith2021implicitSGD}. When $\tilde f$
is conservative, i.e. $\tilde f = -\nabla \tilde L$, gradient descent can be
interpreted as minimizing a modified loss $\tilde L$ rather than $L$.

\paragraph{Known results in standard deep learning.}
For the empirical risk
$L(w) = \tfrac{1}{N}\sum_v \ell(f(w,x_v),y_v)$ 
defined over i.i.d.\ samples, backward error analysis shows that, up to terms of order \(O(\epsilon^3)\), full-batch Gradient Descent (GD) and mini-batch Stochastic Gradient Descent (SGD) respectively optimize the following modified losses~\citep{Barrett2021implicitGD}:
\begin{equation}
\label{eq:gd-loss}
\tilde L_{\mathrm{GD}}(w)
= L(w) + \tfrac{\epsilon}{4}\,\|\nabla L(w)\|^2,
\end{equation}
\begin{equation}
\label{eq:sgd-loss}
\tilde L_{\mathrm{SGD}}(w)
= L(w) + \tfrac{\epsilon}{4}\,\|\nabla L(w)\|^2
+ \tfrac{\epsilon}{4m}\sum_{k=0}^{m-1}\|\nabla \hat L_k(w) - \nabla L(w)\|^2,
\end{equation}
where $\hat L_k$ denotes the loss on mini-batch $k$. The first correction
penalizes large gradients and favors flatter regions of the loss
landscape, which generally correspond to parameters with better generalization performance~\citep{hochreiter1997flat}. The second reflects gradient variance
across batches and vanishes in the full-batch limit.

\paragraph{Full-graph GNN training.}
We extend the backward analysis of \citet{Barrett2021implicitGD} from i.i.d. empirical-risk minimization to full-graph GNN training. The modified-loss construction is agnostic to the source of coupling in the loss, relying only on the additivity of the training objective and gradient-descent dynamics, both of which are preserved under GNN training. Thus, backward analysis applies directly to gradient descent on $L_{\mathrm{full}}$. Eq.~\ref{eq:full_loss}
yields the modified objective derived in Appendix~\ref{app:full_training}:
\begin{equation}
\label{eq:full-graph-gnn-loss}
\tilde L_{\mathrm{full}}(w)
= L_{\mathrm{full}}(w)
+ \tfrac{\epsilon}{4}\,\big\|\nabla L_{\mathrm{full}}(w)\big\|^2.
\end{equation}
The functional form matches the i.i.d.\ result, but the geometry of
$L_{\mathrm{full}}$ is shaped by the graph: node losses are coupled through the
topology and the receptive fields of the GNN.

\paragraph{Sampled-subgraph mini-batch training.}
The same argument applies to any mini-batch GNN training scheme based on
subgraph sampling: the resulting modified loss does not depend on the specific
sampling rule used to construct each batch, only on the per-batch losses it
induces. We therefore state the result generically, denoting by
$\hat L_{\mathrm{Sampl}}^{B_k}$ the loss computed on batch $B_k$ obtained from
an arbitrary sampler $\mathrm{Sampl}$ (e.g., RNS, neighbor sampling,
Cluster-GCN, GraphSAINT). Processing $m$ batches
$\{B_0,\dots,B_{m-1}\}$ once constitutes one epoch, with update
\begin{equation}
w_{t+m} = w_t - \epsilon \sum_{k=0}^{m-1}\nabla \hat L_{\mathrm{Sampl}}^{B_k}(w_{t+k}).
\end{equation}
Define the epoch-averaged partial loss
$\bar L_{\mathrm{Sampl}}(w) = \tfrac{1}{m}\sum_{k=0}^{m-1}\hat L_{\mathrm{Sampl}}^{B_k}(w)$.
In expectation over the batch order in the epoch, the effective modified objective becomes
(Appendix~\ref{app:sampl_training})
\begin{equation}
\label{eq:mini-batch-gnn-loss}
\tilde L_{\mathrm{Sampl}}(w)
= \bar L_{\mathrm{Sampl}}(w)
+ \tfrac{\epsilon}{4}\,\big\|\nabla \bar L_{\mathrm{Sampl}}(w)\big\|^2
+ \tfrac{\epsilon}{4} \cdot \underbrace{\tfrac{1}{m}\sum_{k=0}^{m-1}
   \big\|\nabla \hat L_{\mathrm{Sampl}}^{B_k}(w)-\nabla \bar L_{\mathrm{Sampl}}(w)\big\|^2}_{\displaystyle R(w)}.
\end{equation}

\paragraph{Interpretation.}
In both cases, a finite step size does not only optimize the nominal loss, it introduces an implicit bias acting as a regularizer. In the full-graph case, the dynamics are biased toward
small-gradient regions of $L_{\mathrm{full}}$, where flatness is defined
through the graph-coupled loss. In the sampled-subgraph case, two effects
compound. First, the optimized objective is itself biased, since in general
$\bar L_{\mathrm{Sampl}} \neq L_{\mathrm{full}}$: training minimizes an
empirical average over sampled subgraphs rather than the full-graph loss.
Second, the additional term $R(w)$ is the variance
of batch gradients, and arises specifically from chaining SGD steps on
distinct batches within an epoch. It favors parameters whose updates are stable under graph
sampling.

Equation~\eqref{eq:mini-batch-gnn-loss} yields two testable predictions. First, removing chained per-batch updates should suppress $R(w)$ while preserving the sampling bias in $\bar L_{\mathrm{Sampl}}$. Second, because the result is sampler-agnostic, different samplers should differ through the bias and variance terms induced by their batches. We test the first prediction in Section~\ref{sec:empirical_eval} and quantify the second across samplers in Section~\ref{subsec:gradient_variance}.

\subsection{Empirical evaluation of the implicit regularization with RNS}
\label{sec:empirical_eval}

\begin{table}[h]
\centering
\caption{\textbf{Test accuracy on \textsc{ogbn-products} (in \%).}
Mean $\pm$ 95\% CI over 5 seeds. Both implicit regularization sources are necessary: sampling bias alone (RNS no batch) does not suffice, and neither does chained updating alone (full graph).}
\begin{tabular}{lccc}
\toprule
\textbf{Training regime} & \textbf{SGD (\%)} & \textbf{SGD+M (\%)} & \textbf{Adam (\%)} \\
\midrule
Full graph
& 72.04 $\pm$ 0.11
& 78.89 $\pm$ 0.13
& 80.08 $\pm$ 0.26 \\
RNS (no batch)
& 70.94 $\pm$ 0.15
& 76.84 $\pm$ 0.40
& 83.01 $\pm$ 0.26 \\
RNS
& \textbf{77.70 $\pm$ 0.29}
& \textbf{82.60 $\pm$ 0.20}
& \textbf{83.15 $\pm$ 0.16} \\

\bottomrule
\end{tabular}

\label{tab:implicit_reg_results}
\end{table}

We test the first prediction by training GraphSAGE on \textsc{ogbn-products} under three regimes:

\begin{itemize}
    \item \textbf{Full graph}: standard full-graph training, implicitly minimizing
    $L_{\mathrm{full}}+\tfrac{\epsilon}{4}\|\nabla L_{\mathrm{full}}\|^2$.

    \item \textbf{RNS (no batch)}: RNS batches are sampled, but their gradients are accumulated into one update, implicitly minimizing
    $\bar L_{\mathrm{RNS}}+\tfrac{\epsilon}{4}\|\nabla \bar L_{\mathrm{RNS}}\|^2$ and removing $R(w)$.

    \item \textbf{RNS}: standard mini-batch training with one update per batch, exposing both the sampling bias and $R(w)$ as in Eq.~\eqref{eq:mini-batch-gnn-loss}.
\end{itemize}

We run each regime with SGD, SGD with momentum (SGD+M), and Adam. Details are in Appendix~\ref{app:implicit-reg-impl-details}.

Table~\ref{tab:implicit_reg_results} reports test accuracy across regimes and optimizers. Replacing $L_{\mathrm{full}}$ with $\bar L_{\mathrm{RNS}}$ changes the optimization problem by removing edges between nodes assigned to different batches, thereby injecting sampling noise into training. With Adam, this perturbation improves accuracy, suggesting that the induced noise acts as a useful regularizer and that Adam's adaptive gradient normalization may help exploit it. In contrast, for SGD and SGD+M, RNS (no batch) underperforms the full-graph baseline, indicating that the same sampling-induced noise can hinder optimization when updates are not adaptively preconditioned. RNS training recovers this loss and exceeds full-graph accuracy across optimizers. This shows that the regularization effect predicted by Eq.~\eqref{eq:mini-batch-gnn-loss} becomes beneficial when training performs chained mini-batch updates. Overall, the results indicate that both sources of implicit regularization, the sampling bias in $\bar L_{\mathrm{RNS}}$ and the chaining effect captured by $R(w)$, are empirically visible in the performance and jointly necessary for the strong performance of RNS. The theory--practice gap for Adam is further discussed in Section~\ref{subsec:theory_gap}.

\subsection{On the particularity of RNS implicit regularization}

\label{subsec:gradient_variance}
\begin{figure}[t]
    \centering

    \begin{subfigure}[t]{0.245\linewidth}
        \centering
        \includegraphics[width=\linewidth]{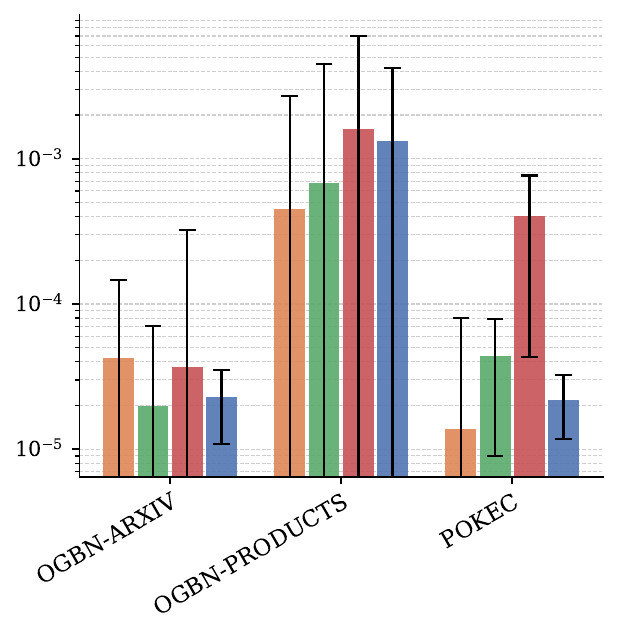}
        \refstepcounter{subfigure}\label{fig:implicit_reg_bias}
        {\scriptsize (\alph{subfigure})~$\big|L_{\mathrm{full}}-\bar L_{\mathrm{Sampl}}\big|$}
    \end{subfigure}
    \hfill
    \begin{subfigure}[t]{0.245\linewidth}
        \centering
        \includegraphics[width=\linewidth]{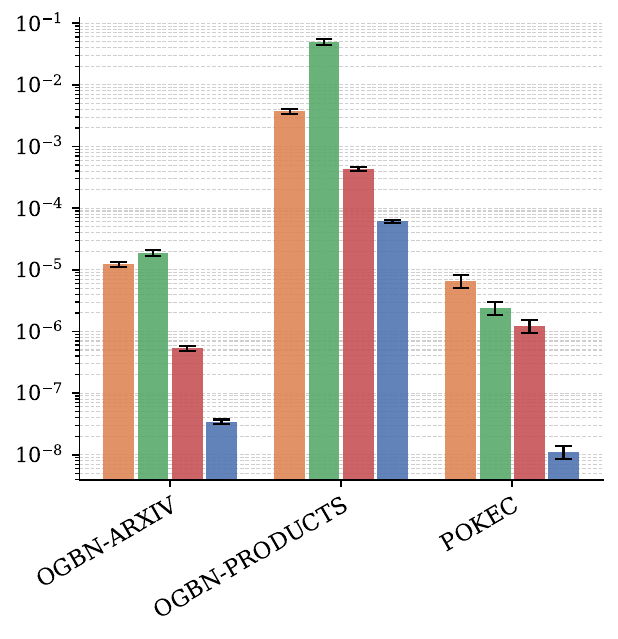}
        \refstepcounter{subfigure}\label{fig:implicit_reg_variance_loss}
        {\scriptsize (\alph{subfigure})~$\operatorname{Var}(\hat L_{\mathrm{Sampl}}^{B_k})$}
    \end{subfigure}
    \hfill
    \begin{subfigure}[t]{0.245\linewidth}
        \centering
        \includegraphics[width=\linewidth]{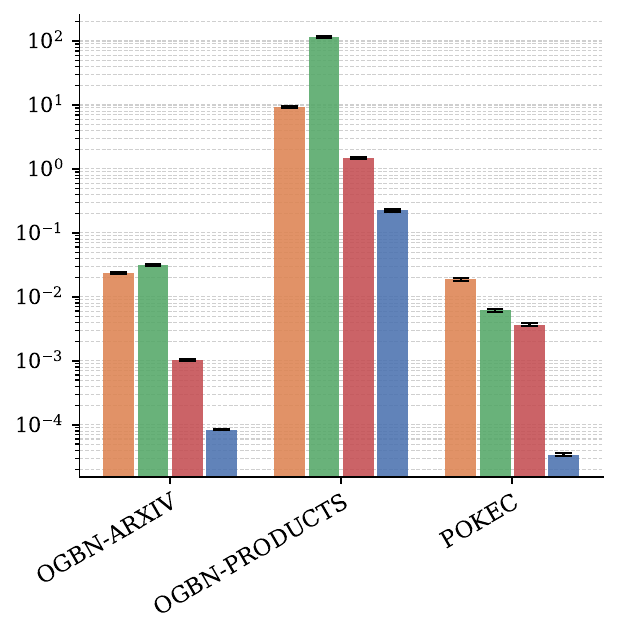}
        \refstepcounter{subfigure}\label{fig:implicit_reg_R}
        {\scriptsize (\alph{subfigure})~$R(w)$}
    \end{subfigure}
    \hfill
    \begin{subfigure}[t]{0.245\linewidth}
        \centering
        \includegraphics[width=\linewidth]{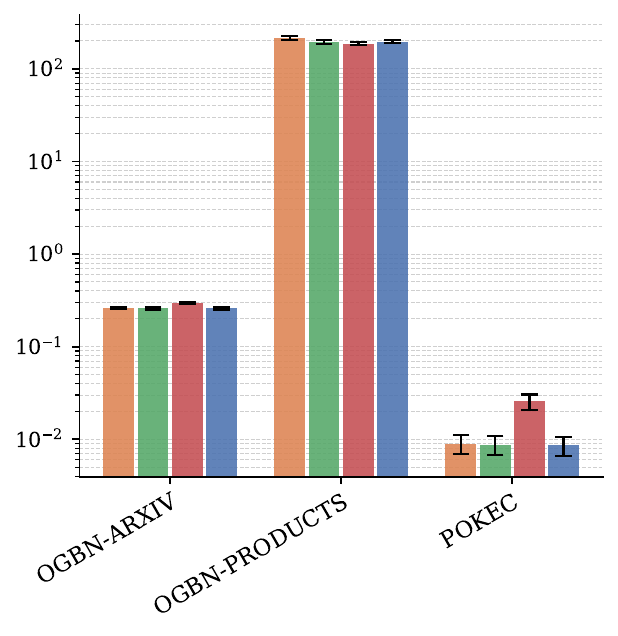}
        \refstepcounter{subfigure}\label{fig:implicit_reg_norm_grad_avg_sq}
        {\scriptsize (\alph{subfigure})~$\|\nabla\bar L_{\mathrm{Sampl}}\|^2$}
    \end{subfigure}

    \includegraphics[width=0.5\linewidth]{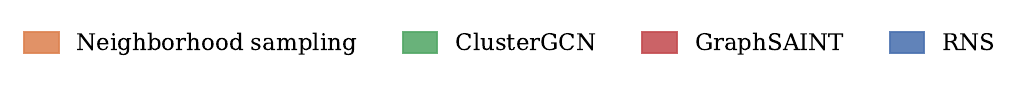}\\[0.5em]

    \caption{Implicit regularization terms induced by different mini-batch samplers at random initialization. Statistics computed from 50 batches. Mean $\pm$ 95\% CI over 100 seeds.}

    \label{fig:implicit_reg_all_metrics}
\end{figure}

We measure the implicit regularization induced by mini-batch samplers at random initialization, before optimization can shape it, using $50$ sampled batches over $100$ random seeds for Neighborhood Sampling, ClusterGCN, GraphSAINT, and RNS.

Figure~\ref{fig:implicit_reg_all_metrics} shows three consistent patterns: all samplers induce a similar loss bias $|L_{\text{full}}-\bar{L}_{\text{Sampl}}|$; per-batch loss variance and gradient variance $R(w)$ differ by orders of magnitude, with RNS lowest in both; and the averaged-objective gradient norm $\|\nabla \bar{L}_{\text{Sampl}}\|^2$ is similar across methods.

These differences reflect how batches are constructed. Appendix~\ref{app:rns-bias} shows 
that, unlike other sampling methods,  the  bias of the RNS objective contains no node-subsampling term and is controlled entirely 
by edge-removal sensitivity, which is small for message-passing GNNs. 
Appendix~\ref{app:degree-thinning} further shows that RNS preserves the power-law degree 
tail in every mini-batch, and Appendix~\ref{app:alpha_R_correlation} shows empirically 
that this structural preservation is strongly correlated with low $R(w)$ across samplers. 
By contrast, ClusterGCN's METIS partitions co-locate community-correlated nodes, making 
clusters systematically dissimilar; Neighborhood Sampling restricts supervision to seed 
nodes, so variance is dominated by the small effective training set; and GraphSAINT biases 
sampling towards high-degree nodes, distorting the degree distribution.

Two consequences follow. First, the implicit objective is sampler-dependent: bias, loss variance, and gradient variance depend strongly on the sampler, so different samplers solve different learning problems. Second, because evaluation uses the full graph, useful batch gradients should approximate those of $L_{\text{full}}$. Among the methods considered, RNS does this most directly, combining relatively low bias, low gradient variance, and unbiased per-batch estimates of full-graph node statistics. In the next section, we empirically benchmark these sampling methods to assess the practical consequences of their different implicit biases.

\section{Experiments and results}

\subsection{Benchmark against other sampling-based training}

\begin{table}[ht]

\centering

\setlength{\tabcolsep}{4pt}

\caption{\textbf{Full-graph vs.\ sampling-based training.}
Accuracy (\%, mean $\pm$ 95\% CI over 5 seeds), time, and peak memory \tablefootnote{All methods are trained for 1000 epochs (2000 for \textsc{pokec}). Neighborhood Sampling for 20, as its CPU-based sampling and small batches make each epoch an order of magnitude slower. (See Appendix~\ref{app:experimental_details}).}. RNS is faster and more accurate, with competitive memory use.}

\begin{tabular}{llcccc}

\hline

Dataset & Method & Accuracy (\%) & Runtime (s) & GPU (GB) & RAM (GB) \\

\hline

\multirow{6}{*}{\textsc{ogbn-arxiv}}

  & Full                  & \textit{72.28 $\pm$ 0.38} & \textit{2m00s} & 3.94  & 1.53 \\
  & Neighborhood sampling & 71.08 $\pm$ 0.47          & 6m49s          & 1.61  & 1.67 \\
  & ClusterGCN            & 71.67 $\pm$ 0.12          & 9m00s          & 1.60  & 1.68 \\
  & SAINT                 & 71.38 $\pm$ 0.09          & 6m00s          & 1.60  & 1.59 \\
  & LADIES                & 68.86 $\pm$ 0.26          & 7m07s          & 2.43  & 1.60 \\
  & RNS                   & \textbf{72.51 $\pm$ 0.19} & \textbf{1m40s} & 2.38  & 1.51 \\

\hline

\multirow{6}{*}{\textsc{ogbn-products}}

  & Full                  & 80.42 $\pm$ 0.50          & \textit{45m50s} & 62.06 & 1.58 \\
  & Neighborhood sampling & 80.56 $\pm$ 0.28          & 3h32m43s        & 24.59 & 10.51 \\
  & ClusterGCN            & 81.51 $\pm$ 0.57          & 55m50s          & 19.89 & 11.55 \\
  & SAINT                 & \textit{82.71 $\pm$ 0.18} & 1h37m30s        & 19.79 & 4.32 \\
  & LADIES                & 78.33 $\pm$ 0.48          & 4h28m40s        & 40.90 & 3.09 \\
  & RNS                   & \textbf{83.11 $\pm$ 0.17} & \textbf{11m30s} & 20.35 & 1.93 \\

\hline

\multirow{6}{*}{\textsc{pokec}}
  & Full                  & \textit{83.40 $\pm$ 0.23} & 37m10s         & 55.43 & 1.63 \\
  & Neighborhood sampling & 80.70 $\pm$ 0.13          & 4h03m15s       & 29.92 & 4.85 \\
  & ClusterGCN            & 80.97 $\pm$ 0.13          & \textbf{27m10s} & 10.69 & 6.55 \\
  & SAINT                 & 78.30 $\pm$ 0.06          & 1h21m50s       & 11.04 & 2.63 \\
  & LADIES                & 56.09 $\pm$ 0.09          & 3h47m38s       & 19.96 & 1.70 \\
  & RNS                   & \textbf{83.98 $\pm$ 0.08} & \textit{29m30s} & 28.66 & 2.17 \\

\hline

\end{tabular}

\label{tab:main-results}

\end{table}

We benchmark on three commonly used large-scale graph datasets (\textsc{ogbn-arxiv}, \textsc{ogbn-products}, and \textsc{pokec}), all using a GraphSAGE backbone. We tune hyperparameters for each dataset and each method separately with Bayesian search. Full implementation details and hyperparameters are reported in Appendix~\ref{sec:implementation_details}. As shown in Table~\ref{tab:main-results}, RNS achieves the best test accuracy on all three datasets while training faster than competing sampling strategies and using comparable GPU memory.

On \textsc{ogbn-products}, all sampling methods perform on par with or slightly above full-graph training, and the differences between samplers are small. On \textsc{ogbn-arxiv} and \textsc{pokec}, however, the baseline samplers tend to fall behind full-graph training, while RNS stays close to or slightly above it. With only three datasets we cannot draw firm conclusions, but one possible explanation relates to the structure of the graphs. \textsc{ogbn-products} is highly homophilic and locally dense, so a sampled subgraph is likely to retain most of the label-relevant signal, which may make topology-aware samplers a reasonable fit for the task. \textsc{ogbn-arxiv} and \textsc{pokec} have lower homophily, so limiting message passing to a sampled subgraph may discard useful longer-range information and introduce bias in the gradients. RNS avoids this issue by construction: it ignores graph structure when forming mini-batches and still propagates on the full graph, which can be viewed as a form of stochastic regularization that does not alter the message-passing computation.

\subsection{Speedup}

We evaluate wall-clock efficiency using validation time-to-accuracy targets.
Across all datasets, targets, and five seeds, both full-graph training and RNS
reach the target accuracies. RNS consistently reduces time-to-target, with
wall-clock speedups ranging from about $2\times$ to $12\times$ depending on the
dataset and whether startup overhead is included. See
Appendix~\ref{app:rns_speedup} for the full decomposition.

The speedup comes from two effects. First, RNS makes epochs cheaper. For a
GraphSAGE model with fixed depth and hidden dimension, the dominant cost is
neighborhood aggregation, which scales approximately with the number of edges
processed. Full-graph training processes all edges each epoch. In contrast, RNS
partitions nodes into $m$ random batches and runs message passing on the induced
subgraphs. A fixed edge appears in a given batch with probability approximately
$1/m^2$, so one batch contains about $|E|/m^2$ edges in expectation, and one RNS
epoch processes about $|E|/m$ induced edges across all batches. Thus the
message-passing cost is reduced by about a factor $m$, although realized epoch
speedups are smaller because of validation, logging, data-loading, optimizer,
and initialization overheads.

Second, RNS changes the epoch-to-update ratio: one RNS epoch contains $m$
optimizer steps, whereas one full-graph epoch contains one. Hence
epochs-to-target speedup should be interpreted as faster progress per epoch,
not necessarily fewer parameter updates. We therefore report both epoch-time
and epochs-to-target speedups in Appendix~\ref{app:rns_speedup}. RNS improves
both across all evaluated targets.

\subsection{Evaluation on GraphLand datasets}
\label{sec:graphland}

\begin{table}[t]
\centering
\caption{\textbf{GraphLand results.}
Full-graph vs.\ RNS performance under random-low (RL) and random-high (RH) regimes. Multiclass: accuracy; binary: average precision (\%). Mean $\pm$ 95\% CI.}
\label{tab:graphland_rns}
\resizebox{\textwidth}{!}{
\begin{tabular}{lcccccc}
\toprule
Dataset
& \multicolumn{3}{c}{RL}
& \multicolumn{3}{c}{RH} \\
\cmidrule(lr){2-4}\cmidrule(lr){5-7}
& Full & RNS & $\Delta$
& Full & RNS & $\Delta$ \\
\midrule
\multicolumn{7}{c}{\textbf{Multiclass classification}} \\
\midrule
\textsc{hm-categories}
& $\mathbf{57.41 \pm 0.34}$ & $57.05 \pm 0.56$ & $-0.36$
& $\mathbf{72.86 \pm 0.34}$ & $72.33 \pm 0.26$ & $-0.53$ \\
\textsc{pokec-regions}
& $37.12 \pm 1.09$ & $\mathbf{41.25 \pm 0.22}$ & $+4.13$
& $38.20 \pm 0.56$ & $\mathbf{47.79 \pm 0.50}$ & $+9.59$ \\
\textsc{web-topics}
& $47.39 \pm 0.06$ & $\mathbf{47.76 \pm 0.08}$ & $+0.37$
& $49.29 \pm 0.11$ & $\mathbf{50.01 \pm 0.05}$ & $+0.72$ \\
\midrule
\multicolumn{7}{c}{\textbf{Binary classification}} \\
\midrule
\textsc{tolokers-2}
& $\mathbf{54.41 \pm 0.19}$ & $54.41 \pm 0.34$ & $-0.00$
& $58.83 \pm 1.01$ & $\mathbf{61.34 \pm 0.44}$ & $+2.51$ \\
\textsc{city-reviews}
& $77.87 \pm 0.09$ & $\mathbf{78.05 \pm 0.05}$ & $+0.18$
& $80.65 \pm 0.08$ & $\mathbf{80.84 \pm 0.21}$ & $+0.19$ \\
\textsc{artnet-exp}
& $42.47 \pm 0.77$ & $\mathbf{42.78 \pm 0.42}$ & $+0.31$
& $48.50 \pm 0.25$ & $\mathbf{49.28 \pm 0.31}$ & $+0.78$ \\
\textsc{web-fraud}
& $\mathbf{12.51 \pm 0.04}$ & $11.44 \pm 0.28$ & $-1.07$
& $\mathbf{20.39 \pm 0.32}$ & $20.29 \pm 0.18$ & $-0.09$ \\
\bottomrule
\end{tabular}
}
\end{table}

To assess whether RNS gains generalize, we further evaluate it on seven node-classification datasets from GraphLand~\cite{bazhenov2024graphland}. These datasets span a wide range of sizes (from $\sim$12K to $\sim$2.9M nodes), densities, homophily levels, and label spaces, and each comes with two official splits: \emph{Random Low} (RL), with a 10\%/10\%/80\% train/validation/test partition, and \emph{Random High} (RH), with a 50\%/25\%/25\% partition, corresponding to scarce and abundant supervision regimes. We closely follow the original GraphLand protocol, with the number of RNS parts treated as an additional hyperparameter in $\{2,3,5,7,10\}$. Full details on the training pipeline and hyperparameter search are given in Appendix~\ref{app:graphland}, and results are reported in Table~\ref{tab:graphland_rns}.

Across the 14 (dataset, regime) configurations, RNS matches or improves on full-graph training in 12 cases. Its gains are larger under abundant supervision (RH) than scarce supervision (RL), where performance is mainly limited by label scarcity. The few failures occur on structurally extreme datasets: very small and dense graphs such as \textsc{hm-categories} and \textsc{tolokers-2}, and the highly imbalanced \textsc{web-fraud} task (with less than 1\% positive labels), where rare predictive patterns may be disrupted. Otherwise, RNS improves performance across varying homophily levels, densities, and graph sizes, supporting its use as a robust drop-in alternative to full-graph training.

\subsection{Does RNS generalize across architectures?}
\label{subsec:rns_architectures}

\begin{table}[t]
\centering
\caption{\textbf{Architecture-wise Full-graph vs.\ RNS comparison.}
Test accuracy (\%, mean $\pm$ 95\% CI over 5 seeds) by dataset and architecture. RNS benefits all message passing architectures. We briefly comment on GAT OOM in Appendix~\ref{subsec:architectures_details}.}
\label{tab:testacc_full_rns_pct_compact}
\resizebox{\linewidth}{!}{
\begin{tabular}{lcccccc}
\toprule
& \multicolumn{2}{c}{\textsc{ogbn-arxiv}} & \multicolumn{2}{c}{\textsc{ogbn-products}} & \multicolumn{2}{c}{\textsc{pokec}} \\
\cmidrule(lr){2-3} \cmidrule(lr){4-5} \cmidrule(lr){6-7}
Model & Full & RNS & Full & RNS & Full & RNS \\
\midrule
GCN
& $73.10 \pm 0.22$ & $\mathbf{73.26 \pm 0.28}$
& $77.55 \pm 0.30$ & $\mathbf{80.81 \pm 0.25}$
& $84.62 \pm 0.05$ & $\mathbf{84.98 \pm 0.12}$ \\
GraphSAGE
& $72.28 \pm 0.38$ & $\mathbf{72.51 \pm 0.19}$
& $80.42 \pm 0.50$ & $\mathbf{83.11 \pm 0.17}$
& $83.40 \pm 0.23$ & $\mathbf{83.98 \pm 0.08}$ \\
GAT
& $72.65 \pm 0.10$ & $\mathbf{72.74 \pm 0.26}$
& \textsc{OOM} & \textsc{OOM}
& \textsc{OOM} & \textsc{OOM} \\
SGFormer
& $\mathbf{72.54 \pm 0.07}$ & $72.42 \pm 0.16$
& $72.42 \pm 0.25$ & $\mathbf{74.64 \pm 0.15}$
& $\mathbf{77.23 \pm 0.13}$ & $75.38 \pm 0.18$ \\
\bottomrule
\end{tabular}
}
\end{table}

Having fixed the backbone in the previous experiments, we now test whether RNS transfers across architectures. We compare full-graph training to RNS for GCN~\citep{kipf2016semi}, GraphSAGE~\citep{hamilton2017inductive}, GAT~\citep{velivckovic2017graph}, and SGFormer~\citep{wu2023sgformer}, performing an independent hyperparameter search per (architecture, dataset, training mode) triple. GAT runs out of memory on \textsc{ogbn-products} and \textsc{pokec} under both regimes. We briefly comment on that in Appendix~\ref{subsec:architectures_details}.

Table~\ref{tab:testacc_full_rns_pct_compact} shows that RNS matches or improves full-graph training for every message-passing backbone on every dataset. The only exception is SGFormer, which underperforms full-graph training on \textsc{ogbn-arxiv} and on \textsc{pokec}. We attribute this to SGFormer's global attention branch, whose all-pairs interactions are truncated when only a random subset of nodes is observed at each step. Overall, RNS is essentially architecture-agnostic for standard message-passing GNNs, while its interaction with global-attention models might require further investigation.

\section{Discussion and limitations}
\label{sec:limitations}

\subsection{From SGD to Adam: a theory-practice gap}
\label{subsec:theory_gap}
Our backward error analysis is derived for vanilla SGD in order to keep the modified-equation expansion tractable and to make the role of sampling-induced perturbations explicit. In practice, Adam consistently achieves higher test accuracy than SGD across all regimes we consider (Table~\ref{tab:implicit_reg_results}), so a natural question is whether the implicit objective of Eq.~\eqref{eq:mini-batch-gnn-loss} extends to adaptive optimizers. Prior work partially closed this gap. \citet{ghosh2023implicit} show that SGD with Heavy-ball momentum, where $\beta$ denotes the momentum coefficient, admits a modified loss whose implicit regularizer is amplified by a factor $\tfrac{1+\beta}{1-\beta}$ relative to SGD. This provides a principled explanation for the generalization gains of momentum and motivates our SGD+M baseline. \citet{Cattaneo2024implicitADAM} derive an ODE approximation for Adam and show the existence of an implicit regularizer that penalizes a perturbed $\ell_1$-norm of the loss gradient, but their analysis is restricted to the full-batch setting. Extending this analysis to stochastic Adam remains, to our knowledge, open, as coordinate-wise adaptive normalization renders the modified ODE nonlinear. Empirically, our experiments suggest that the implicit regularization mechanism identified for SGD persists qualitatively under Adam, but extending the formal analysis is left for future work.

\subsection{Impact of number of batches per epoch}

\begin{figure}[t]
  \centering
  \includegraphics[width=\textwidth]{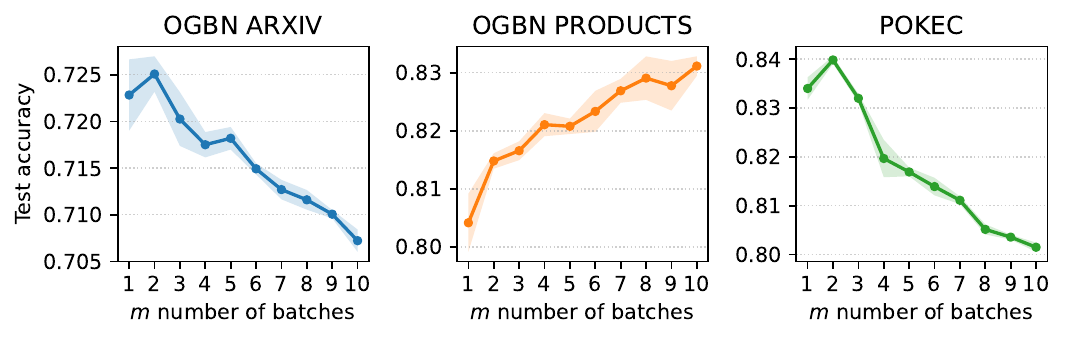}
  \caption{\textbf{Impact of the number of batches $m$.}
Test accuracy vs.\ $m$ (batches per epoch) on \textsc{ogbn-arxiv}, \textsc{ogbn-products}, and \textsc{pokec}, shaded areas indicate 95\% CI over 5 seeds. We fix the architecture and tune training hyperparameters separately for each value of $m$} 
  \label{fig:impact-m}
\end{figure}

Figure~\ref{fig:impact-m} shows that test accuracy is highly sensitive to the number of batches per epoch $m$, even when hyperparameters are selected independently for each value of $m$ and the architecture is fixed. Thus, $m$ should be treated as an important hyperparameter. This limitation is not specific to RNS: other graph sampling methods typically introduce two or more sampling parameters, such as fanout, walk length, batch size, or sampling depth, which also require tuning. In practice, RNS remains comparatively simple to tune; we recommend starting from $m=2$ and increasing $m$ until accuracy plateaus.

To understand the structural drivers of these trends, Figure~\ref{fig:sampling-properties} tracks properties of the sampled subgraphs as $m$ varies. Sparsity-related metrics such as edge count, average degree, and connected-component structure evolve similarly across the three datasets and do not explain the differences in accuracy behavior. Subgraph diameter is more informative: it increases substantially with $m$ on \textsc{ogbn-arxiv} and \textsc{pokec}, but remains stable on \textsc{ogbn-products}, correlating with the accuracy degradation observed on the first two datasets. Our analysis is restricted to integer-valued disjoint partitions into $m$ batches; sampling nodes with replacement at a fixed fraction per step would allow continuous control over subgraph density and may perform better at intermediate scales. Finally, the largest gains occur on the two largest graphs, \textsc{ogbn-products} and \textsc{pokec}, suggesting that improvements may grow with graph size.

\section{Conclusion}

RNS provides a simple yet effective alternative to sophisticated GNN mini-batching methods. Our analysis shows that sampling does not merely approximate full-graph training: it optimizes a modified implicit objective with a sampled loss term and a batch gradient variance penalty. Compared with other sampling methods, this batch gradient variance regularization term is several orders of magnitude smaller, making training more stable by producing sampled subgraphs that remain closer to the full graph.

Empirically, RNS delivers a strong compute--accuracy trade-off on large transductive benchmarks. It generally achieves better test performance than full-graph training while reducing memory usage. Overall, our results establish RNS as both a practical baseline for scalable GNN training and a theoretically grounded form of implicit regularization.

\newpage

% \begin{ack}
% Use unnumbered first level headings for the acknowledgments. All acknowledgments
% go at the end of the paper before the list of references. Moreover, you are required to declare
% funding (financial activities supporting the submitted work) and competing interests (related financial activities outside the submitted work).
% More information about this disclosure can be found at: \url{https://neurips.cc/Conferences/2026/PaperInformation/FundingDisclosure}.

% Do {\bf not} include this section in the anonymized submission, only in the final paper. You can use the \texttt{ack} environment provided in the style file to automatically hide this section in the anonymized submission.
% \end{ack}

\bibliographystyle{plainnat}
\bibliography{neurips_2026}

%%%%%%%%%%%%%%%%%%%%%%%%%%%%%%%%%%%%%%%%%%%%%%%%%%%%%%%%%%%%

\newpage

\appendix

\section{Implicit objective derivation via backward analysis}
\label{app:implicit_proof}

Our proofs closely follow the framework and several key arguments of \citet{Smith2021implicitSGD}. We refer readers to that work for additional background and intuition.

\subsection{Full graph training \label{app:full_training}}

\paragraph{Setup and assumptions.}
We consider gradient descent with step size $\alpha>0$,
\begin{equation}
w_{t+1}=w_t+\alpha f(w_t),
\end{equation}
where $f:\mathbb{R}^d\to\mathbb{R}^d$ is the training vector field (typically $f=-\nabla L$). Backward analysis approximates this discrete update by the flow of a modified ODE
\begin{equation}
\dot w=\tilde f(w),
\end{equation}
whose vector field admits the expansion in $\epsilon$
\begin{equation}\label{eq:f_hat}
\tilde f(w)=f(w)+\epsilon g(w)+O(\epsilon^2),
\end{equation}
with $\epsilon$ proportional to $\alpha$. $g(w)$ represents the contribution to the modified flow of the small finite learning rate $\alpha$. 

If the network uses ReLU, then $f$ is piecewise smooth: its derivatives exist except on a measure-zero set of activation-switching parameters.

\paragraph{Local regularity and boundedness.}
We consider a finite time horizon and assume the iterates (and the interpolating trajectory) remain in a compact set $K\subset\mathbb{R}^d$ that does not intersect the nonsmooth set. On $K$ we assume $\tilde f$ is $C^2$. In particular, $\tilde f$ is Lipschitz on $K$, and by compactness there exists $C>0$ such that for all $w\in K$,
\begin{equation}
\|\tilde f(w)\|\le C,\qquad \|\nabla \tilde f(w)\|\le C,\qquad \|\nabla^2\tilde f(w)\|\le C.
\end{equation}

\paragraph{Trajectory deviation bound.}
Iterating the update rule yields, for any $k\ge1$,
\begin{equation}
w_{t+k}-w_t
=
\alpha
\sum_{j=0}^{k-1}
\tilde f(w_{t+j}),
\end{equation}
and therefore
\begin{equation}
\|w_{t+k}-w_t\|
\le
\alpha
\sum_{j=0}^{k-1}
\|\tilde f(w_{t+j})\|
\le
C\,\alpha\,k.
\end{equation}

\paragraph{Second-order Taylor expansion.}
For each $k\ge0$, we expand $\tilde f(w_{t+k})$ around $w_t$:
\begin{equation}
\begin{aligned}
\tilde f(w_{t+k})
&=
\tilde f(w_t)
+
\nabla \tilde f(w_t)
\bigl(w_{t+k}-w_t\bigr)
+
R_k ,
\end{aligned}
\end{equation}
where
\begin{equation}
\|R_k\|
\le
\frac{1}{2}
\|\nabla^2 \tilde f(w_t)\|
\|w_{t+k}-w_t\|^2
\le
\frac{C^3}{2}\alpha^2 k^2.
\end{equation}

\paragraph{Summation of $m$ updates.}
We write
\begin{equation}
w_{t+m}
=
w_t
+
\alpha
\sum_{k=0}^{m-1}
\tilde f(w_{t+k}).
\end{equation}
Using the Taylor expansion,
\begin{equation}
\begin{aligned}
\sum_{k=0}^{m-1}
\tilde f(w_{t+k})
=
m\,\tilde f(w_t)
+
\nabla \tilde f(w_t)
\sum_{k=0}^{m-1}
(w_{t+k}-w_t)
+
\sum_{k=0}^{m-1}
R_k .
\end{aligned}
\end{equation}

The remainder satisfies
\begin{equation}
\left\|
\sum_{k=0}^{m-1}
R_k
\right\|
\le
\frac{C^3}{2}
\alpha^2
\sum_{k=0}^{m-1}
k^2
=
O(\alpha^2 m^3).
\end{equation}

\paragraph{Evaluation of the inner sum.}
We control
$\sum_{k=0}^{m-1}(w_{t+k}-w_t)$.
Using the trajectory representation,
\begin{equation}
\sum_{k=0}^{m-1}
(w_{t+k}-w_t)
=
\alpha
\sum_{k=0}^{m-1}
\sum_{j=0}^{k-1}
\tilde f(w_{t+j})
=
\alpha
\sum_{j=0}^{m-2}
(m-1-j)\tilde f(w_{t+j}).
\end{equation}

Expanding $\tilde f(w_{t+j})$ around $w_t$,
\begin{equation}
\tilde f(w_{t+j})
=
\tilde f(w_t)
+
R_j',
\end{equation}
with
\begin{equation}
\|R_j'\|
\le
\|\nabla \tilde f(w_t)\|
\|w_{t+j}-w_t\|
\le
C^2 \alpha j.
\end{equation}

Substituting,
\begin{equation}
\begin{aligned}
\sum_{k=0}^{m-1}
(w_{t+k}-w_t)
&=
\alpha
\sum_{j=0}^{m-2}
(m-1-j)\tilde f(w_t)
\\
&\quad
+
\alpha
\sum_{j=0}^{m-2}
(m-1-j)R_j'.
\end{aligned}
\end{equation}

The first term is
\begin{equation}
\alpha
\sum_{j=0}^{m-2}
(m-1-j)
=
\alpha
\frac{m(m-1)}{2}.
\end{equation}

The latter satisfies
\begin{equation}
\left\|
\alpha
\sum_{j=0}^{m-2}
(m-1-j)R_j'
\right\|
\le
\alpha^2 C^2
\sum_{j=0}^{m-2}
j(m-1-j)
=
O(\alpha^2 m^3).
\end{equation}

Thus,
\begin{equation}
\sum_{k=0}^{m-1}
(w_{t+k}-w_t)
=
\alpha
\frac{m(m-1)}{2}
\tilde f(w_t)
+
O(\alpha^2 m^3).
\end{equation}

\paragraph{Final discrete expansion.}
Substituting back,
\begin{equation}
w_{t+m}
=
w_t
+
m\alpha
\tilde f(w_t)
+
\alpha^2
\frac{m(m-1)}{2}
\nabla \tilde f(w_t)
\tilde f(w_t)
+
O(\alpha^3 m^3).
\end{equation}

\paragraph{Continuous-time limit and implicit regularization.}
Let $\alpha=\epsilon/m$. Then
\begin{equation}
\begin{aligned}
w(t+\epsilon)
&=
w(t)
+
\epsilon \tilde f(w)
+
\frac{\epsilon^2}{2}
\nabla \tilde f(w)\tilde f(w)
+
O(\epsilon^3)
\\
&=
w(t)
+
\epsilon f(w)
+
\epsilon^2
\left(
g(w)
+
\frac{1}{2}
\nabla f(w)f(w)
\right)
+
O(\epsilon^3),
\end{aligned}
\end{equation}
using \eqref{eq:f_hat}. We choose $\alpha=\epsilon/m$ so that the cumulative step size after $m$ discrete updates is the fixed increment $\epsilon$, allowing us to compare the $m$-step discrete map directly to the continuous-time expansion over time $\epsilon$.

Matching the Euler discretization
\begin{equation}
w_{k+1}
=
w_k
+
\epsilon f(w_k),
\end{equation}
yields
\begin{equation}
g(w)
=
-\frac{1}{2}
\nabla f(w)f(w).
\end{equation}

For $f(w)=-\nabla L_{\mathrm{full}}(w)$,
\begin{equation}
\begin{aligned}
g(w)
&=
-\frac{1}{2}
\nabla^2 L_{\mathrm{full}}(w)
\nabla L_{\mathrm{full}}(w)
\\
&=
-\frac{1}{4}
\nabla
\|\nabla L_{\mathrm{full}}(w)\|^2.
\end{aligned}
\end{equation}

Therefore, gradient descent with step size $\epsilon$ follows, up to order $O(\epsilon^3)$,
\begin{equation}
\dot w
=
-\nabla \tilde L_{\mathrm{full}}(w),
\end{equation}
with modified loss
\begin{equation}
\tilde L_{\mathrm{full}}(w)
=
L_{\mathrm{full}}(w)
+
\frac{\epsilon}{4}
\|\nabla L_{\mathrm{full}}(w)\|^2.
\end{equation}

\subsection{Mini-batch GNN training under subgraph sampling}
\label{app:sampl_training}

\paragraph{Setup and notation.}
Let $N=|V|$ and fix a batch size $B$, with $N$ divisible by $B$ and
$m=N/B$. Let $\{B_0,\dots,B_{m-1}\}$ denote a partition of $V$ produced by a
generic subgraph sampler $\mathrm{Sampl}$ (e.g., RNS, neighbor sampling,
Cluster-GCN, GraphSAINT). At each epoch, the batches are processed in
a uniformly random order. The derivation below holds for any such sampler:
only the per-batch loss $\hat L^{B_k}_{\mathrm{Sampl}}$ depends on the choice
of $\mathrm{Sampl}$.

For batch $B_k$, define the partial loss
\begin{equation}
\hat L^{B_k}_{\mathrm{Sampl}}(w)
=
\frac{1}{|B_k^\mathrm{train}|}
\sum_{v\in B^{\mathrm{train}}_k}
\ell\!\big(f_w(X_{B_k},E_{B_k})[v],\,y_v\big),
\end{equation}
and the associated vector field
\begin{equation}
f_k(w)
:=
-\nabla \hat L^{B_k}_{\mathrm{Sampl}}(w).
\end{equation}

The SGD update over one epoch is
\begin{equation}
w_{t+m}
=
w_t
+
\alpha
\sum_{k=0}^{m-1}
f_k(w_{t+k}).
\end{equation}

We assume there exists $C>0$ such that for all $k$ and $w$,
\begin{equation}
\|f_k(w)\| \le C,
\qquad
\|\nabla f_k(w)\| \le C,
\qquad
\|\nabla^2 f_k(w)\| \le C.
\end{equation}

\paragraph{Trajectory deviation bound.}
For any $k\ge1$,
\begin{equation}
w_{t+k}-w_t
=
\alpha
\sum_{l=0}^{k-1}
f_l(w_{t+l}),
\end{equation}
hence
\begin{equation}
\|w_{t+k}-w_t\|
\le
C\,\alpha\,k.
\end{equation}

\paragraph{Taylor expansion.}
For each $k$,
\begin{equation}
\begin{aligned}
f_k(w_{t+k})
&=
f_k(w_t)
+
\nabla f_k(w_t)
\bigl(w_{t+k}-w_t\bigr)
+
R_k ,
\end{aligned}
\end{equation}
with
\begin{equation}
\|R_k\|
\le
\frac{1}{2}
\|\nabla^2 f_k\|
\|w_{t+k}-w_t\|^2
\le
\frac{C^3}{2}\alpha^2 k^2.
\end{equation}

Summing over $k$,
\begin{equation}
\begin{aligned}
\sum_{k=0}^{m-1}
f_k(w_{t+k})
&=
\sum_{k=0}^{m-1}
f_k(w_t)
+
\sum_{k=0}^{m-1}
\nabla f_k(w_t)
\bigl(w_{t+k}-w_t\bigr)
+
O(\alpha^2 m^3).
\end{aligned}
\end{equation}

\paragraph{Interaction term.}
Using
\begin{equation}
w_{t+k}-w_t
=
\alpha
\sum_{l=0}^{k-1}
f_l(w_t)
+
O(\alpha^2 m^2),
\end{equation}
we obtain
\begin{equation}
\begin{aligned}
\sum_{k=0}^{m-1}
\nabla f_k(w_t)
\bigl(w_{t+k}-w_t\bigr)
&=
\alpha
\sum_{k=0}^{m-1}
\sum_{l<k}
\nabla f_k(w_t) f_l(w_t)
+
O(\alpha^2 m^3).
\end{aligned}
\end{equation}

Define
\begin{equation}
\xi(w)
:=
\sum_{k=0}^{m-1}
\sum_{l<k}
\nabla f_k(w) f_l(w).
\end{equation}

\paragraph{Expectation over batch order.}
Since we are interested in the bias introduced by SGD, we condition on the
batch contents and take expectation over the random permutation:
\begin{equation}
\mathbb{E}[\xi(w)]
=
\frac{1}{2}
\sum_{k\neq l}
\nabla f_k(w) f_l(w).
\end{equation}

Note that in general, the sum for $l<k$ is not equal to the sum for $l>k$.
This reflects the fact that, algebraically, the learned weights depend on the
mini-batch ordering.

Equivalently,
\begin{equation}
\begin{aligned}
\mathbb{E}[\xi(w)]
&=
\frac{1}{2}
\left(
\sum_{k=0}^{m-1}
\nabla f_k(w)
\right)
\left(
\sum_{l=0}^{m-1}
f_l(w)
\right)
-
\frac{1}{2}
\sum_{k=0}^{m-1}
\nabla f_k(w) f_k(w).
\end{aligned}
\end{equation}

Since, for all $f$,
\begin{equation}
\nabla f(w) f(w)
=
\frac{1}{2}
\nabla
\|f(w)\|^2,
\end{equation}
we obtain
\begin{equation}
\begin{aligned}
\mathbb{E}[\xi(w)]
&=
\frac{1}{4}
\nabla
\left\|
\sum_{k=0}^{m-1}
f_k(w)
\right\|^2
-
\frac{1}{4}
\nabla
\sum_{k=0}^{m-1}
\|f_k(w)\|^2 .
\end{aligned}
\end{equation}

\paragraph{Effective modified flow.}
Define the mean vector field
\begin{equation}
f(w)
=
\frac{1}{m}
\sum_{k=0}^{m-1}
f_k(w).
\end{equation}

Matching with the second-order continuous-time expansion yields
\begin{equation}
g(w)
=
-\frac{1}{4m^2}
\nabla
\sum_{k=0}^{m-1}
\|f_k(w)\|^2.
\end{equation}

Note that during the identification, we also set $\epsilon = m\alpha$.

Therefore, mini-batch GNN training under sampler $\mathrm{Sampl}$ follows
\begin{equation}
\dot w
=
-\nabla \tilde L_{\mathrm{Sampl}}(w),
\end{equation}
with modified loss
\begin{equation}
\begin{aligned}
\tilde L_{\mathrm{Sampl}}(w)
&=
\frac{1}{m}
\sum_{k=0}^{m-1}
\hat L^{B_k}_{\mathrm{Sampl}}(w)
+
\frac{\epsilon}{4m}
\sum_{k=0}^{m-1}
\left\|
\nabla
\hat L^{B_k}_{\mathrm{Sampl}}(w)
\right\|^2 .
\end{aligned}
\end{equation}

\paragraph{Decomposition of the implicit regularization term.}
The additional regularization induced by mini-batch sampling is
\begin{equation}
\frac{\epsilon}{4m}
\sum_{k=0}^{m-1}
\left\|
\nabla \hat L^{B_k}_{\mathrm{Sampl}}(w)
\right\|^2.
\end{equation}

Introduce the averaged partial loss
\begin{equation}
\bar L_{\mathrm{Sampl}}(w)
:=
\frac{1}{m}
\sum_{k=0}^{m-1}
\hat L^{B_k}_{\mathrm{Sampl}}(w),
\end{equation}
whose gradient satisfies
\begin{equation}
\nabla \bar L_{\mathrm{Sampl}}(w)
=
\frac{1}{m}
\sum_{k=0}^{m-1}
\nabla \hat L^{B_k}_{\mathrm{Sampl}}(w).
\end{equation}

Using the variance decomposition,
\begin{equation}
\begin{aligned}
\frac{1}{m}
\sum_{k=0}^{m-1}
\left\|
\nabla \hat L^{B_k}_{\mathrm{Sampl}}(w)
\right\|^2
&=
\left\|
\nabla \bar L_{\mathrm{Sampl}}(w)
\right\|^2
+
\frac{1}{m}
\sum_{k=0}^{m-1}
\left\|
\nabla \hat L^{B_k}_{\mathrm{Sampl}}(w)
-
\nabla \bar L_{\mathrm{Sampl}}(w)
\right\|^2 .
\end{aligned}
\end{equation}

The modified loss therefore admits the decomposition
\begin{equation}
\begin{aligned}
\tilde L_{\mathrm{Sampl}}(w)
&=
\bar L_{\mathrm{Sampl}}(w)
+
\frac{\epsilon}{4}
\left\|
\nabla \bar L_{\mathrm{Sampl}}(w)
\right\|^2
+
\frac{\epsilon}{4m}
\sum_{k=0}^{m-1}
\left\|
\nabla \hat L^{B_k}_{\mathrm{Sampl}}(w)
-
\nabla \bar L_{\mathrm{Sampl}}(w)
\right\|^2 .
\end{aligned}
\end{equation}

\paragraph{Discussion about the approximation.}
In classical deep learning settings, the precise meaning of ``small finite'' learning rates $\epsilon$ in backward analysis is subtle, we refer to the extended discussion in~\cite{Smith2021implicitSGD}. Briefly, this regime allows $O(m^{2}\epsilon^{2})$ terms to be potentially significant while still neglecting $O(m^{3}\epsilon^{3})$ terms, which effectively requires $m\epsilon$ to remain small enough for the truncation to be valid. In our setting this requirement is less problematic because the number of mini-batches per epoch stays small: in all experiments we have $m \le 15$, since larger $m$ would excessively damage the topology of the sampled subgraphs.

\section{Bias of RNS}
\label{app:rns-bias}

We provide an upper bound on the bias of the RNS implicit objective with respect to the
full-graph loss $L_{\mathrm{full}}$. The argument shows that the bias has
\emph{no} contribution from node subsampling, and is entirely controlled by the
GNN's stability under edge removal.

\paragraph{Setting.}
Let $|V|=N$ and $|V^{\mathrm{train}}| = N_t$, and assume $m$ divides  $N$. RNS samples a
uniformly random partition $\pi = (B_0, \dots, B_{m-1})$ of $V$ into $m$ ordered
blocks of size $N/m$. For convenience, define the per-node losses
\[
\ell_v^{\mathrm{full}}(w) := \ell\!\bigl(f(w, X, E)[v],\, y_v\bigr),
\qquad
\ell_v^{B}(w) := \ell\!\bigl(f(w, X_B, E_B)[v],\, y_v\bigr) \text{ for } v \in B,
\]
so that, for a generic block $B$,
\[
\hat L_{\mathrm{RNS}}^{B}(w)
=
\frac{1}{|B^{\mathrm{train}}|}\sum_{v \in B^{\mathrm{train}}} \ell_v^{B}(w),
\qquad
L_{\mathrm{full}}(w)
=
\frac{1}{N_t}\sum_{v \in V^{\mathrm{train}}} \ell_v^{\mathrm{full}}(w),
\]
where $B^{\mathrm{train}} := B \cap V^{\mathrm{train}}$.
We work conditionally on the event $\mathcal{E} = \{|B^{\mathrm{train}}| \geq 1\}$,
whose complement satisfies
\[
\Pr(\mathcal{E}^c)
= \frac{\binom{N - N_t}{N/m}}{\binom{N}{N/m}}
\quad\text{(0 if $N/m > N - N_t$)},
\]
which is exponentially small whenever $N_t / m \gg 1$.

\paragraph{Lemma 1: marginal law of a single block.}
Under a uniformly random ordered partition $\pi$ of $V$ into $m$ blocks of
size $N/m$, the marginal distribution of any fixed block $B_k$ is uniform over
$\bigl\{S \subseteq V : |S| = N/m\bigr\}$.
%Indeed, for any such $S$, the number of partitions of $V$ in which $B_k = S$ equals the number of ordered partitions of $V \setminus S$ into $m-1$ blocks of size $N/m$, namely $(N - N/m)!/((N/m)!)^{m-1}$. This count does not depend on $S$, so $\Pr(B_k = S)$ is constant in $S$.

\paragraph{Lemma 2: conditional law of the supervised intersection.}
Let $B$ denote any  of the blocks. Let $t\ge 1$. Conditionally on $|B^{\mathrm{train}}| = t$, the
set $B^{\mathrm{train}}$ is uniformly distributed over $t$-subsets of
$V^{\mathrm{train}}$. By Lemma~1, $B$ is uniform over $(N/m)$-subsets of $V$,
so for any fixed $t$-subset $S \subseteq V^{\mathrm{train}}$,
\[
\Pr(B^{\mathrm{train}} = S)
= \frac{\binom{N - N_t}{\,N/m - t\,}}{\binom{N}{\,N/m\,}},
\]
which depends only on $|S| = t$, not on $S$ itself. Conditioning on
$|B^{\mathrm{train}}| = t$ therefore preserves uniformity. As an immediate
consequence, for every $v \in V^{\mathrm{train}}$ and every $t \geq 1$,
\begin{equation}
\label{eq:rns-marginal-prob}
\Pr\!\bigl(v \in B \,\big|\, |B^{\mathrm{train}}| = t\bigr)
= \Pr\!\bigl(v \in B^{\mathrm{train}} \,\big|\, |B^{\mathrm{train}}| = t\bigr)
= \frac{t}{N_t}.
\end{equation}

\paragraph{Main result.}
Assume the per-node loss is $C_\ell$-Lipschitz in its first argument:
$|\ell(z_1, y) - \ell(z_2, y)| \leq C_\ell\,\|z_1 - z_2\|$ for all $z_1, z_2, y$.
Define the GNN edge-removal sensitivity
\begin{equation}
\label{eq:rns-sigma}
\sigma(w)
\;:=\;
\max_{v \in V^{\mathrm{train}}}\;
\sup_{t \geq 1}\;
\mathbb{E}_\pi\!\left[\,
\bigl\|f(w, X_B, E_B)[v] - f(w, X, E)[v]\bigr\|
\;\Big|\; v \in B,\, |B^{\mathrm{train}}| = t
\,\right].
\end{equation}
%Note that $\sigma(w) = 0$ whenever the receptive field of every supervised node is contained in its batch (e.g.\ $m = 1$). 
We show that
\begin{equation}
\label{eq:rns-bias-bound}
\bigl|\,\mathbb{E}_\pi[\hat L_{\mathrm{RNS}}^{B}(w) \mid \mathcal{E}] - L_{\mathrm{full}}(w)\,\bigr|
\;\leq\; C_\ell\, \sigma(w).
\end{equation}
The same bound holds for the epoch-averaged objective
$\bar L_{\mathrm{RNS}}(w) := \tfrac{1}{m}\sum_{k=0}^{m-1}\hat L_{\mathrm{RNS}}^{B_k}(w)$
under the analogous event
$\bar{\mathcal{E}} := \bigcap_{k}\{|B_k^{\mathrm{train}}| \geq 1\}$,
since the $m$ blocks share the marginal law of Lemma~1 and the bound is linear.

\paragraph{Proof outline.}
Introduce the auxiliary quantity
\[
\tilde L^{B}(w)
:= \frac{1}{|B^{\mathrm{train}}|}\sum_{v \in B^{\mathrm{train}}} \ell_v^{\mathrm{full}}(w),
\]
which evaluates the loss on the same supervised nodes as
$\hat L_{\mathrm{RNS}}^{B}$, but using the \emph{full-graph} forward pass. We
decompose
\begin{equation}
\label{eq:rns-decomp}
\hat L_{\mathrm{RNS}}^{B} - L_{\mathrm{full}}
\;=\; \underbrace{\bigl(\tilde L^{B} - L_{\mathrm{full}}\bigr)}_{\text{(I): node subsampling}}
\;+\; \underbrace{\bigl(\hat L_{\mathrm{RNS}}^{B} - \tilde L^{B}\bigr)}_{\text{(II): edge removal}},
\end{equation}
and show that $\mathbb{E}[\text{(I)} \mid \mathcal{E}] = 0$ and
$\bigl|\mathbb{E}[\text{(II)} \mid \mathcal{E}]\bigr| \leq C_\ell \sigma(w)$.

\paragraph{Step 1: term (I) has zero conditional expectation.}
Fix $t \geq 1$. By linearity of expectation and \eqref{eq:rns-marginal-prob},
\[
\mathbb{E}\!\left[\sum_{v \in B^{\mathrm{train}}} \ell_v^{\mathrm{full}}(w) \,\bigg|\, |B^{\mathrm{train}}| = t\right]
= \sum_{v \in V^{\mathrm{train}}} \ell_v^{\mathrm{full}}(w)\,
  \Pr\!\bigl(v \in B \mid |B^{\mathrm{train}}| = t\bigr)
= \frac{t}{N_t} \sum_{v \in V^{\mathrm{train}}} \ell_v^{\mathrm{full}}(w),
\]
hence
\[
\mathbb{E}\!\left[\tilde L^{B}(w) \,\bigg|\, |B^{\mathrm{train}}| = t\right]
= \frac{1}{t}\cdot\frac{t}{N_t}\sum_{v \in V^{\mathrm{train}}} \ell_v^{\mathrm{full}}(w)
= L_{\mathrm{full}}(w).
\]
This identity holds for every $t \geq 1$; averaging over the conditional law of
$|B^{\mathrm{train}}|$ given $\mathcal{E}$ yields
\[
\mathbb{E}\!\left[\tilde L^{B}(w) - L_{\mathrm{full}}(w) \,\big|\, \mathcal{E}\right]
= 0.
\]
This step uses only Lemma~2 and is the formal statement that the
node-subsampling component of RNS is unbiased.

\paragraph{Step 2: pointwise bound on term (II).}
For each $v \in B$, define
$\delta_v(B, w) := \|f(w, X_B, E_B)[v] - f(w, X, E)[v]\|$.
By the Lipschitz assumption,
$|\ell_v^{B}(w) - \ell_v^{\mathrm{full}}(w)| \leq C_\ell\, \delta_v(B, w)$, hence
\begin{equation}
\label{eq:rns-pointwise}
\bigl|\hat L_{\mathrm{RNS}}^{B}(w) - \tilde L^{B}(w)\bigr|
\;\leq\; \frac{C_\ell}{|B^{\mathrm{train}}|}\sum_{v \in B^{\mathrm{train}}} \delta_v(B, w).
\end{equation}

\paragraph{Step 3: expected bound on term (II).}
Fix $t \geq 1$. Writing the sum over $B^{\mathrm{train}}$ as an indicator sum
over $V^{\mathrm{train}}$ and using \eqref{eq:rns-marginal-prob},
\begin{align*}
\mathbb{E}\!\left[\sum_{v \in B^{\mathrm{train}}} \delta_v(B, w) \,\bigg|\, |B^{\mathrm{train}}| = t\right]
&= \sum_{v \in V^{\mathrm{train}}}
   \mathbb{E}\!\left[\mathbf{1}[v \in B]\,\delta_v(B, w) \,\big|\, |B^{\mathrm{train}}| = t\right] \\
&= \sum_{v \in V^{\mathrm{train}}}
   \Pr\!\bigl(v \in B \mid |B^{\mathrm{train}}| = t\bigr)\,
   \mathbb{E}\!\left[\delta_v(B, w) \mid v \in B,\, |B^{\mathrm{train}}| = t\right] \\
&= \frac{t}{N_t} \sum_{v \in V^{\mathrm{train}}}
   \mathbb{E}\!\left[\delta_v(B, w) \mid v \in B,\, |B^{\mathrm{train}}| = t\right].
\end{align*}
Dividing by $t$,
\[
\mathbb{E}\!\left[\frac{1}{|B^{\mathrm{train}}|}\sum_{v \in B^{\mathrm{train}}} \delta_v(B, w) \,\bigg|\, |B^{\mathrm{train}}| = t\right]
= \frac{1}{N_t}\sum_{v \in V^{\mathrm{train}}}
  \mathbb{E}\!\left[\delta_v(B, w) \mid v \in B,\, |B^{\mathrm{train}}| = t\right]
\;\leq\; \sigma(w),
\]
where the last inequality follows directly from the definition
\eqref{eq:rns-sigma}: each summand is bounded by $\sigma(w)$ uniformly in $t$,
and the average over $v$ is bounded by the maximum. Since the bound holds for
every $t \geq 1$, averaging over the conditional law of $|B^{\mathrm{train}}|$
given $\mathcal{E}$ preserves it:
\[
\mathbb{E}\!\left[\frac{1}{|B^{\mathrm{train}}|}\sum_{v \in B^{\mathrm{train}}} \delta_v(B, w) \,\bigg|\, \mathcal{E}\right]
\;\leq\; \sigma(w).
\]
Combining with \eqref{eq:rns-pointwise} and Jensen's inequality
($|\mathbb{E}[X]| \leq \mathbb{E}[|X|]$),
\[
\bigl|\mathbb{E}[\hat L_{\mathrm{RNS}}^{B} - \tilde L^{B} \mid \mathcal{E}]\bigr|
\;\leq\;
\mathbb{E}\!\left[\bigl|\hat L_{\mathrm{RNS}}^{B} - \tilde L^{B}\bigr| \,\big|\, \mathcal{E}\right]
\;\leq\; C_\ell\,\sigma(w).
\]

\paragraph{Step 4: combine.}
Applying the triangle inequality to \eqref{eq:rns-decomp} and combining
Steps~1 and~3,
\[
\bigl|\mathbb{E}[\hat L_{\mathrm{RNS}}^{B} \mid \mathcal{E}] - L_{\mathrm{full}}\bigr|
\;\leq\; \bigl|\mathbb{E}[\tilde L^{B} - L_{\mathrm{full}} \mid \mathcal{E}]\bigr|
       + \bigl|\mathbb{E}[\hat L_{\mathrm{RNS}}^{B} - \tilde L^{B} \mid \mathcal{E}]\bigr|
\;\leq\; 0 + C_\ell\,\sigma(w),
\]
which establishes \eqref{eq:rns-bias-bound}.

\paragraph{Extension to the epoch-averaged objective.}
By linearity, $\mathbb{E}[\bar L_{\mathrm{RNS}}\mid\bar{\mathcal{E}}]
= \tfrac{1}{m}\sum_k \mathbb{E}[\hat L_{\mathrm{RNS}}^{B_k}\mid\bar{\mathcal{E}}]$.
Lemma~1 (and a symmetry argument under $\bar{\mathcal{E}}$, which is invariant
under permutation of block indices) ensures that each term has the same value
as $\mathbb{E}[\hat L_{\mathrm{RNS}}^{B_0}\mid\bar{\mathcal{E}}]$, and the
bound \eqref{eq:rns-bias-bound} carries over with $\mathcal{E}$ replaced by
$\bar{\mathcal{E}}$. By a union bound,
$\Pr(\bar{\mathcal{E}}^c)\leq m\Pr(\mathcal{E}^c)$, which remains exponentially
small under $N_t/m \gg 1$.

\paragraph{Intuition: why $\sigma(w)$ scales with $1 - 1/m$.}
By Lemma~1, $B$ is a uniformly random $(N/m)$-subset of $V$, so for any
pair of distinct nodes $u, w$, the probability that both lie in $B$ is
\[
\Pr(u \in B,\, w \in B)
\;=\; \frac{(N/m)(N/m - 1)}{N(N - 1)}
\;=\; \frac{1}{m^2}\cdot\frac{N - m}{N - 1}.
\]
Equivalently, an edge $(u, w)$ is \emph{cut} (i.e.\ removed in $E_B$)
with probability
\[
\Pr\!\bigl((u, w) \notin E_B\bigr)
\;=\; 1 - \frac{1}{m^2}\cdot\frac{N-m}{N-1}
\;=\; \left(1 - \frac{1}{m}\right)\!\left(1 + O\!\bigl(\tfrac{1}{m}\bigr)\right).
\]
Conditioning on $v \in B$ shifts these probabilities by at most $O(1/N)$.
Hence, if $\mathcal{E}_L(v)$ denotes the set of edges in the $L$-hop
receptive field of $v$, the expected number of cut edges given $v \in B$
is
\[
\mathbb{E}\!\bigl[\#\{e \in \mathcal{E}_L(v) : e \notin E_B\}\,\big|\, v \in B\bigr]
\;\leq\; \bigl|\mathcal{E}_L(v)\bigr|\cdot\left(1 - \frac{1}{m}\right) + O(1).
\]
For any GNN whose output at $v$ is $C_f$-Lipschitz with respect to edge
removals in $\mathcal{E}_L(v)$, this immediately gives
\[
\sigma(w)
\;\leq\; C_f\,\bar{\mathcal{E}}_L\!\left(1 - \frac{1}{m}\right) + O(1/N),
\qquad
\bar{\mathcal{E}}_L := \max_v |\mathcal{E}_L(v)|.
\]
Observations: 
\begin{itemize}
    \item $\sigma(w) = 0$ at $m = 1$ (full-graph training)
    \item $\sigma(w)$ scales linearly in $1 - 1/m$, so it is small in the regime $m = O(1)$ that we use in practice but becomes larger quickly when we increase $m$
    \item the prefactor
$C_f \bar{\mathcal{E}}_L$ is architecture- and graph-dependent. It is small
for normalised message-passing GNNs (GCN, SAGE, GAT) on real graphs,
but unbounded for global attention, where every node attends to every
other and removing any edge perturbs every output.
\end{itemize}

\paragraph{Why the proof fails for structure-based samplers.}
Step~1 is the only step that uses uniformity, through Lemma~2. For ClusterGCN
and GraphSAINT, Lemma~2 fails: clusters are not exchangeable, and random-walk
samplers have stationary distribution proportional to degree, so conditional on
$|B^{\mathrm{train}}| = t$ the supervised intersection $B^{\mathrm{train}}$ is
\emph{not} uniform over $t$-subsets of $V^{\mathrm{train}}$. Consequently
$\Pr(v \in B \mid |B^{\mathrm{train}}| = t) \neq t/N_t$ in general, and the
cancellation in Step~1 breaks. The bias of these samplers therefore contains a
non-vanishing first-order term \emph{in addition to} the edge-removal term,
which cannot be reduced by tuning $m$. Only RNS isolates the objective bias to
the architecture-dependent quantity $\sigma(w)$.

\section{Tail degree distribution under RNS}
\label{app:degree-thinning}

\paragraph{Setting.}
Let $G=(V,E)$ be a graph on $N$ nodes, and let
\[
\nu(d)=\frac{\#\{v\in V:\deg(v)=d\}}{N}
\]
be its empirical degree distribution. We assume that the degree distribution
has a power-law tail: there exist constants $\alpha>1$ and $C>0$ such that
\[
\nu(d)=C d^{-\alpha}(1+o(1)),
\qquad d\to\infty.
\]
Power-law degree distributions have been widely observed in real-world
networks, including technological, social, and biological networks
\cite{barabasi1999emergence}.

Partition $V$ uniformly at random into $m$ batches of equal size $N/m$.
Let $U$ be a node chosen uniformly from $V$, independently of the
partition. Write $B$ for the batch containing $U$,
\[
D=\deg_G(U),\qquad
\widetilde D=\deg_{G[B]}(U),
\]
and set $q=1/m$.

\paragraph{Proposition.}
For every sequence $k=k(N)\to\infty$ such that $k=o(N/m)$, one has
\[
\widetilde p(k):=\mathbb P(\widetilde D=k)
=Cq^{\alpha-1}k^{-\alpha}(1+o(1))
=Cm^{1-\alpha}k^{-\alpha}(1+o(1)).
\]
In other words, the power-law tail of the degree distribution is preserved
on RNS-subsampled graphs.

\paragraph{Proof.}
Since $U$ is chosen uniformly from $V$, $\mathbb P(D=d)=\nu(d)$.
Conditional on $D=d$, the number $\widetilde D$ of neighbors of $U$ falling
in the same batch as $U$ has the hypergeometric law
\[
H_{N,d}(k):=\mathbb P(\widetilde D=k\mid D=d)
=\frac{\binom{d}{k}\binom{N-1-d}{N/m-1-k}}{\binom{N-1}{N/m-1}},
\]
so that
\[
\widetilde p(k)=\sum_{d\ge k}\nu(d)\,H_{N,d}(k).
\]
We analyze this sum by treating $H_{N,d}(k)$ (up to normalization) as a
probability mass function in $d$ that concentrates at $d\approx k/q$, and
combining this with the asymptotic homogeneity of $\nu$ on the
concentration window.

Throughout, set
\[
M=N-1,\qquad n=N/m-1,
\]
so that $M+1=N$ and $n+1=N/m=qN$.

\medskip\noindent
\textbf{Step 1: Total mass of the kernel.}
By Vandermonde's identity,
\[
\sum_{d}\binom{d}{k}\binom{M-d}{n-k}=\binom{M+1}{n+1}=\binom{N}{N/m}.
\]
Hence
\[
\sum_d H_{N,d}(k)
=\frac{\binom{N}{N/m}}{\binom{N-1}{N/m-1}}
=\frac{N}{N/m}=m=\frac1q,
\]
and therefore $\mathsf P_k(d):=qH_{N,d}(k)$ is a probability mass function
in $d$.

\medskip\noindent
\textbf{Step 2: Identification with an order statistic; mean and variance.}
We identify $\mathsf P_k$ with the law of an order statistic.
For a uniformly chosen $r$-subset $\{X_1<\cdots<X_r\}$ of $\{1,\dots,L\}$,
\[
\mathbb P(X_{k+1}=y)
=\frac{\binom{y-1}{k}\binom{L-y}{r-k-1}}{\binom{L}{r}}.
\]
Setting $L=M+1=N$, $r=n+1=N/m$, and $y=d+1$ gives
\[
\mathbb P(X_{k+1}=d+1)
=\frac{\binom{d}{k}\binom{M-d}{n-k}}{\binom{M+1}{n+1}}
=\frac{n+1}{M+1}\,H_{N,d}(k)
=q\,H_{N,d}(k)
=\mathsf P_k(d+1\!-\!1).
\]
Thus, under $\mathsf P_k$, $d$ is distributed as $X_{k+1}-1$, where
$X_{k+1}$ is the $(k{+}1)$-th order statistic of a uniform $(N/m)$-subset
of $\{1,\dots,N\}$.

The standard formulas for such order statistics give
\[
\mathbb E_k[d]
=\frac{(k+1)(L+1)}{r+1}-1
=\frac{(k+1)(N+1)}{N/m+1}-1,
\]
\[
\operatorname{Var}_k(d)
=\frac{(k+1)(r-k)(L+1)(L-r)}{(r+1)^2(r+2)}
=\frac{(k+1)(n-k)(N+1)(N-N/m)}{(N/m+1)^2(N/m+2)}.
\]
In the regime $k=o(N/m)$, this simplifies to
\[
\mathbb E_k[d]=\frac{k}{q}+O(1),
\qquad
\operatorname{Var}_k(d)=O(k),
\]
uniformly in $k$.

\medskip\noindent
\textbf{Step 3: Concentration of the kernel.}
Fix $\varepsilon\in(0,1/2)$ and define
\[
W_k=\Big\{d\ge k:\ \big|d-k/q\big|\le k^{1/2+\varepsilon}\Big\}.
\]
Since $\mathbb E_k[d]=k/q+O(1)$ and $\operatorname{Var}_k(d)=O(k)$,
Chebyshev's inequality gives, for $k$ large enough,
\[
\sum_{d\notin W_k}\mathsf P_k(d)
=\mathbb P_k\!\left(|d-k/q|>k^{1/2+\varepsilon}\right)
=O(k^{-2\varepsilon}).
\]
Since $q$ is fixed,
\[
\sum_{d\notin W_k}H_{N,d}(k)=O(k^{-2\varepsilon}),
\qquad
\sum_{d\in W_k}H_{N,d}(k)=\frac1q+o(1).
\]
Note that for $k$ large enough, $W_k\subset[k,N-1]$: indeed
$k/q-k^{1/2+\varepsilon}>k$ since $q<1$ and $\varepsilon<1/2$, while
$k/q+k^{1/2+\varepsilon}<N-1$ because $k=o(N/m)$.

\medskip\noindent
\textbf{Step 4: Uniform use of the power-law tail on $W_k$.}
By assumption there exist $d_0$ and $C'>0$ such that
$\nu(d)\le C'd^{-\alpha}$ for all $d\ge d_0$. Since $k\to\infty$,
eventually $k\ge d_0$, and then for every $d\ge k$,
\[
\nu(d)\le C'd^{-\alpha}\le C'k^{-\alpha}. \tag{$\ast$}
\]
For $d\in W_k$,
\[
\frac{d}{k/q}=1+O\!\left(qk^{-1/2+\varepsilon}\right)
=1+o(1),
\]
uniformly in $d\in W_k$, so
\[
d^{-\alpha}=\left(\tfrac{k}{q}\right)^{-\alpha}(1+o(1))
\]
uniformly. Combined with $\nu(d)=Cd^{-\alpha}(1+o(1))$ and the fact that
$d\to\infty$ uniformly on $W_k$, this yields
\[
\nu(d)=C\!\left(\tfrac{k}{q}\right)^{-\alpha}(1+o(1)),
\qquad\text{uniformly for }d\in W_k.
\]

\medskip\noindent
\textbf{Step 5: Decomposition.}
Split
\[
\widetilde p(k)
=\underbrace{\sum_{d\in W_k}\nu(d)H_{N,d}(k)}_{=:S_{\mathrm{in}}(k)}
+\underbrace{\sum_{d\notin W_k}\nu(d)H_{N,d}(k)}_{=:S_{\mathrm{out}}(k)}.
\]

\emph{Inner contribution.} Using the uniform estimate from Step 4,
\[
S_{\mathrm{in}}(k)
=C\!\left(\tfrac{k}{q}\right)^{-\alpha}(1+o(1))
\sum_{d\in W_k}H_{N,d}(k)
=\frac{C}{q}\!\left(\tfrac{k}{q}\right)^{-\alpha}(1+o(1)).
\]

\emph{Outer contribution.} By $(\ast)$ and the concentration bound,
\[
S_{\mathrm{out}}(k)
\le C'k^{-\alpha}\sum_{d\notin W_k}H_{N,d}(k)
=O\!\left(k^{-\alpha-2\varepsilon}\right)
=o(k^{-\alpha}).
\]

\medskip\noindent
\textbf{Step 6: Conclusion.}
Combining,
\[
\widetilde p(k)
=\frac{C}{q}\!\left(\tfrac{k}{q}\right)^{-\alpha}(1+o(1))+o(k^{-\alpha})
=Cq^{\alpha-1}k^{-\alpha}(1+o(1)).
\]
Since $q=1/m$, this is equivalently
\[
\widetilde p(k)=Cm^{1-\alpha}k^{-\alpha}(1+o(1)),
\]
proving the proposition.

\section{Datasets}
\label{app:datasets}

We evaluate our method on a diverse collection of large-scale transductive node-classification benchmarks. Our main evaluation uses three widely adopted large graphs: \textsc{ogbn-arxiv}, \textsc{ogbn-products}~\cite{hu2020open}, and \textsc{Pokec}~\cite{takac2012data}. These datasets are standard in the scalable graph learning literature and cover complementary regimes: \textsc{ogbn-arxiv} is a comparatively small large-scale citation graph, \textsc{ogbn-products} is substantially larger while remaining computationally tractable, and \textsc{Pokec} is a large social network with different structural and feature properties from the two OGB datasets.

We further extend our evaluation with seven node-classification datasets from GraphLand~\cite{bazhenov2024graphland}. These benchmarks increase the diversity of graph sizes, label spaces, homophily levels, and structural properties considered in our experiments. We include three multi-class classification datasets, \textsc{hm-categories}, \textsc{pokec-regions}, and \textsc{web-topics}, and four binary classification datasets, \textsc{tolokers-2}, \textsc{city-reviews}, \textsc{artnet-exp}, and \textsc{web-fraud}. We refer readers to the original GraphLand paper for details on the construction of these datasets. Unless otherwise stated, all datasets are processed as undirected graphs. Summary statistics are reported in Table~\ref{tab:datasets}.

\paragraph{\textsc{ogbn-arxiv}.}
\textsc{ogbn-arxiv} is a directed citation network of Computer Science arXiv papers indexed by MAG~\cite{hu2020open}. Nodes correspond to papers, and directed edges represent citation relationships. Each node is associated with a 128-dimensional feature vector obtained by averaging word embeddings from the paper title and abstract. We use the official OGB time-based split, where papers published up to 2017 are used for training, papers published in 2018 for validation, and papers published from 2019 onward for testing.

\paragraph{\textsc{ogbn-products}.}
\textsc{ogbn-products} is an undirected, unweighted Amazon co-purchasing graph~\cite{hu2020open}. Nodes represent products, and edges connect products that are frequently bought together. Node features are constructed from bag-of-words representations of product descriptions and reduced to 100 dimensions using PCA. We use the official sales-rank split, where the top 8\% most popular products are used for training, the next 2\% for validation, and the remaining products for testing. This split reflects a realistic setting in which labels are first collected for popular items.

\paragraph{\textsc{Pokec}.}
\textsc{Pokec} is a large-scale social network from Slovakia~\cite{takac2012data}. We consider the binary node-classification task of predicting user gender. Following \citet{luo2024classic}, we use a random 1/1/8 train/validation/test split. Since the exact split used by \citet{luo2024classic} is not publicly released, we resample the split at the beginning of each training run.

\paragraph{GraphLand datasets.}
GraphLand~\cite{bazhenov2024graphland} provides a collection of large-scale node-classification datasets spanning diverse graph structures, label spaces, and homophily levels. We use the seven GraphLand node-classification datasets listed in Table~\ref{tab:datasets}. Each dataset is provided with two official random split regimes, Random Low (RL) and Random High (RH), which differ in the amount of labeled data available for training. In the RL regime, nodes are randomly partitioned into train/validation/test sets with proportions 10\%/10\%/80\%, whereas in the RH regime the corresponding proportions are 50\%/25\%/25\%. We report results on both regimes.

\begin{table}[t]
  \centering
    \caption{Graph statistics for the datasets used in our experiments. Hom. denotes homophily. Diameter (Diam.) and average shortest-path distance (Avg.\ dist.) are respectively the maximum and the average shortest-path length between two nodes in the same connected component, estimated using BFS sampling. Edges are counted twice since all datasets are processed as undirected graphs.}
  \small
  \setlength{\tabcolsep}{2.6pt}
  \begin{tabular}{l r r r r r r r r r}
    \toprule
    Dataset & Nodes & Edges & Density & Avg.\ deg. & \# Isol. & \# CC & Hom. & Diam. & Avg.\ dist. \\
    \midrule
    \textsc{ogbn-arxiv}     & 169K   & 1.17M & $4.10{\times}10^{-5}$ & 13.8  & 0   & 1   & 0.65 & 24 & 6.12 \\
    \textsc{ogbn-products}  & 2.45M  & 61.9M & $1.03{\times}10^{-5}$ & 50.2  & 48K & 52K & 0.81 & 29 & 5.35 \\
    \textsc{Pokec}          & 1.63M  & 30.6M & $1.15{\times}10^{-5}$ & 27.3  & 0   & 1   & 0.42 & 14 & 4.68 \\
    \midrule
    \textsc{hm-categories}  & 46.5K  & 10.7M & $4.95{\times}10^{-3}$ & 460.9 & 0   & 1   & 0.38 & 13 & 2.45 \\
    \textsc{pokec-regions}  & 1.63M  & 30.6M & $1.15{\times}10^{-5}$ & 27.3  & 0   & 1   & 0.98 & 14 & 4.68 \\
    \textsc{web-topics}     & 2.9M   & 12.4M & $1.47{\times}10^{-6}$ & 8.6   & 0   & 1   & 0.55 & 36 & 3.08 \\
    \textsc{tolokers-2}     & 11.8K  & 2.33M & $1.67{\times}10^{-2}$ & 88.3  & 0   & 1   & 0.10 & 11 & 2.79 \\
    \textsc{city-reviews}   & 148.8K & 1.2M  & $5.42{\times}10^{-5}$ & 15.7  & 0   & 1   & 0.69 & 19 & 4.91 \\
    \textsc{artnet-exp}     & 50.4K  & 560K  & $2.21{\times}10^{-4}$ & 11.1  & 0   & 1   & 0.28 & 13 & 4.42 \\
    \textsc{web-fraud}      & 2.9M   & 12.9M & $1.53{\times}10^{-6}$ & 8.6   & 0   & 1   & 0.32 & 36 & 3.08 \\
    \bottomrule
  \end{tabular}

  \label{tab:datasets}
\end{table}

\section{Experimental Details}
\label{app:experimental_details}

This section describes the experimental protocols used throughout the paper. We provide implementation and evaluation details for each group of experiments, organized into four subsections: the sampling benchmark, the GraphLand benchmark, the ablation study on RNS, and the architecture benchmark.

\subsection{Sampling benchmark}
\label{sec:implementation_details}

This section describes the experimental protocol used to benchmark sampling strategies on large-scale graphs. We consider three datasets, \textsc{ogbn-arxiv}, \textsc{ogbn-products}, and \textsc{pokec}, and compare several mini-batch training schemes under a common architecture and tuning protocol. Our goal is to isolate the effect of the \emph{sampling strategy} itself, while ensuring that each method is evaluated under competitive optimization settings.

Unless otherwise stated, validation and test metrics are computed on the full graph. The only exception is neighbor sampling, for which we also use sampled neighborhoods at evaluation time, following the standard inference pipeline of that method.

The hyperparameter tuning phase was run on a heterogeneous GPU cluster containing NVIDIA H100 GPUs with 80GB of memory, L40S and A40 GPUs with 48GB of memory, and A100 GPUs with 40GB of memory. After tuning, all final training runs used for reporting results were run on H100 GPUs.

\subsubsection{Implementation of sampling strategies}
\label{subsec:sampling_strategies}

All sampling procedures are implemented in PyTorch Geometric (PyG)~\cite{fey2019fast, fey2025scalable}. For all methods except Random Node Sampling (RNS), we use the corresponding PyG data loaders. In all cases, the sampled mini-batch adjacency is converted to a sparse representation before being passed to the model.

\paragraph{Neighbor sampling~\cite{hamilton2017inductive}.}
We use the standard neighbor sampler, parameterized by a fanout vector
$$
\mathbf{f} = (f_1, \dots, f_L),
$$
where $f_\ell$ is the maximum number of neighbors sampled at layer $\ell$, together with a batch size of target nodes. Each mini-batch is formed by first selecting a set of seed nodes, then recursively sampling up to $f_\ell$ neighbors per node. One epoch corresponds to iterating over the full set of seed nodes once.

\paragraph{ClusterGCN~\cite{chiang2019cluster}.}
ClusterGCN is parameterized by the number of graph partitions $C$ and the number of sampled partitions per mini-batch $B$. We first partition the graph into $C$ clusters using METIS~\cite{karypis1998fast}. At each optimization step, we sample $B$ clusters uniformly at random and construct the mini-batch as the induced subgraph over the union of their nodes.

\paragraph{GraphSAINT~\cite{zeng2019graphsaint}.}
GraphSAINT is parameterized by the random-walk length $L_{\mathrm{w}}$ and the number of seed nodes $S$. For each mini-batch, we sample $S$ seeds uniformly at random and launch one random walk of length $L_{\mathrm{w}}$ from each seed. The mini-batch is the induced subgraph over the union of visited nodes.

To make GraphSAINT comparable to RNS in terms of sampled graph volume per epoch, we fix the number of optimization steps per epoch to
$$
\left\lfloor \frac{|V|}{S \cdot L_{\mathrm{w}}} \right\rfloor,
$$
so that the total number of sampled node visits per epoch is on the order of $|V|$, up to overlaps among walks.

\paragraph{LADIES~\cite{zou2019layer}.}
We reimplement LADIES from the original GitHub repository, with a GPU implementation of the sampling procedure. LADIES is parameterized by a batch size of target nodes and a layer-wise sampling budget vector
$$
\mathbf{s} = (s_1, \dots, s_L),
$$
where $s_\ell$ denotes the number of nodes sampled for layer $\ell$. For each mini-batch, LADIES first selects a set of target nodes of size given by the batch size, then samples nodes layer by layer using an importance distribution defined over the one-hop neighbors of the current upper-layer node set. Let $\mathcal{S}_{\ell+1}$ denote the sampled nodes at layer $\ell+1$, and let $\mathbf{A}_{:,\mathcal{S}_{\ell+1}}$ be the submatrix of the normalized adjacency matrix restricted to the columns indexed by $\mathcal{S}_{\ell+1}$. LADIES assigns each candidate node $i$ the importance score
$$
q_i
=
\left\|
\mathbf{A}_{i,\mathcal{S}_{\ell+1}}
\right\|_2^2,
$$
and samples $s_\ell$ nodes with probabilities
$$
p_i
=
\frac{q_i}{\sum_j q_j}.
$$
Thus, nodes with larger aggregate normalized connectivity to the current upper-layer nodes are more likely to be selected. One epoch corresponds to iterating over the full set of target nodes once.

\paragraph{Random Node Sampling (RNS).}
We implement RNS with a custom GPU-based data loader. Since RNS only requires uniformly sampling nodes and extracting the corresponding induced subgraphs, a GPU implementation is substantially faster in our setting than the existing CPU-based PyG routine. RNS is parameterized by the number of partitions $P$, which controls the mini-batch size.

At the beginning of each epoch, we draw a random permutation of all nodes and split it into $P$ disjoint blocks of size $\lfloor |V|/P \rfloor$. Each block defines one mini-batch through its induced subgraph. Pseudocode is provided in Algorithm~\ref{alg:rns}.

\begin{algorithm}[t]
\caption{RNS data loading for one epoch}
\label{alg:rns}
\begin{algorithmic}[1]
\Require Graph $G=(V,E)$, device, number of partitions $P$
\State $G \gets G.\texttt{to}(\texttt{device})$
\State $n \gets |V|$
\State $b \gets \lfloor n / P \rfloor$
\State $\pi \gets \texttt{torch.randperm}(n, \texttt{device}=\texttt{device})$
\For{$i = 0,1,\ldots,P-1$}
    \State $s \gets i \cdot b$
    \State $t \gets s + b$
    \State $S_i \gets \pi[s:t]$
    \State $G_i \gets \texttt{extract\_subgraph\_from\_nodes}(G, S_i)$
    \State $G_i \gets \texttt{to\_sparse\_tensor}(G_i)$
    \State \textbf{yield} $G_i$
\EndFor
\end{algorithmic}
\end{algorithm}

\subsubsection{Fair hyperparameter tuning protocol}
\label{subsec:hpo_protocol}

A central difficulty when comparing sampling methods is that performance depends not only on the sampler itself, but also on the interaction between the sampler, the optimizer, and regularization. To ensure a fair comparison, we use the following protocol.

First, we \emph{fix the model architecture within each dataset} across all sampling methods. In particular, hidden dimension, number of layers, normalization type, input/output projection choices, and residual connections are taken from \citet{luo2024classic}. The resulting architectures are summarized in Table~\ref{tab:arch_hparams}. By keeping the architecture fixed, we reduce the risk that differences in accuracy are driven by model design rather than by sampling.

Second, for \emph{each sampling method and each choice of sampling hyperparameters}, we independently tune the \emph{training-related hyperparameters}. Concretely, for every dataset and every candidate sampling configuration, we run a separate hyperparameter sweep over learning rate, dropout, and weight decay. This is important: we do \emph{not} tune one global optimizer configuration per method and reuse it across all sampling settings. Instead, each sampling configuration is allowed to operate under its own best optimization hyperparameters.

Third, the \emph{sampling hyperparameters themselves} are selected from method-specific search spaces motivated by the original papers. For each candidate sampling configuration, we perform 50 Bayesian-optimization runs with \textsc{Weights \& Biases} sweeps over the training hyperparameters. We then report, for each dataset and method, the configuration with the best validation performance.

This procedure separates the roles of the different hyperparameter classes:
\begin{itemize}
    \item \textbf{Architecture hyperparameters} are fixed per dataset and shared across methods.
    \item \textbf{Sampling hyperparameters} are explored within a method-specific search space.
    \item \textbf{Training hyperparameters} are tuned independently for every candidate sampling configuration.
\end{itemize}
This layered protocol is designed to make the comparison as fair as possible across methods with very different sampling behaviors.

\paragraph{Training hyperparameters.}
Across all datasets and methods, we tune the same set of optimization hyperparameters:
\begin{itemize}
    \item dropout rate,
    \item learning rate,
    \item weight decay.
\end{itemize}
The common search space is:
\begin{center}
\begin{tabular}{ll}
\toprule
\textbf{Hyperparameter} & \textbf{Search space} \\
\midrule
Dropout & $\{0.0, 0.2, 0.3, 0.5, 0.7\}$ \\
Learning rate & $\{10^{-2}, 5\times 10^{-3}, 10^{-3}, 5\times 10^{-4}, 10^{-4}\}$ \\
Weight decay & $\{0, 10^{-4}, 5\times 10^{-4}, 10^{-3}, 5\times 10^{-3}\}$ \\
\bottomrule
\end{tabular}
\end{center}

\paragraph{Architecture hyperparameters.}
We use GraphSAGE as the backbone model for all experiments. The dataset-specific architectural choices inherited from \citet{luo2024classic} are reported in Table~\ref{tab:arch_hparams}. For LADIES, to follow the original paper, we employ a GCN architecture with the same hyperparameters except the number of layers which was fixed 5. 

\begin{table}[t]
\centering
\small
\setlength{\tabcolsep}{3.5pt}
\caption{Architecture hyperparameters used for each dataset. These parameters are fixed across sampling methods. In proj. and out proj. denote input and output projections, respectively. Note that Neighborhood sampling is trained for 20 epochs.}
\label{tab:arch_hparams}
\begin{tabular}{lccccccc}
\toprule
Dataset & Hid. & Layers & In proj. & Out proj. & Res. & Norm & Epochs \\
\midrule
\textsc{ogbn-arxiv}    & 256 & 4 & No & Yes & Yes & BN & 1000 \\
\textsc{ogbn-products} & 256 & 5 & No & No  & No  & LN & 1000 \\
\textsc{pokec}         & 256 & 7 & No & Yes & Yes & BN & 2000 \\
\bottomrule
\end{tabular}
\end{table}

For neighbor sampling, we train for 20 epochs only. This choice accounts for its substantially larger number of mini-batches per epoch, resulting from the smaller batch size, and its considerably higher training cost relative to the other methods (see Appendix \ref{sec:neighbor_sampling_depth}).

\paragraph{Sampling hyperparameter search spaces.}

We now describe the search spaces used for the method-specific sampling hyperparameters. For ClusterGCN and GraphSAINT, we adopt candidate configurations directly from the original papers~\cite{chiang2019cluster, zeng2019graphsaint}. For neighbor sampling, the original paper explores only relatively shallow fanout schedules~\cite{hamilton2017inductive}, so we use standard fanout values and additionally include one deeper configuration to match the depth of our models. For RNS, since there is no canonical search space in prior work, we vary the number of partitions over a broad range.

For neighbor sampling, we evaluate the fanout vectors
$$
\mathbf{f} \in \{[30,10], [30,10,10], [30,10,10,10]\}.
$$
The first two configurations are standard shallow fanout schedules. We additionally consider the depth-4 configuration $[30,10,10,10]$ in order to study the effect of increased sampling depth on predictive performance. We do not consider deeper schedules because, in our implementation, PyG's neighbor sampler runs on the CPU, making the sampling cost too high beyond depth 4 (See Section \ref{sec:neighbor_sampling_depth}).

For ClusterGCN, following the configurations considered in the original paper~\cite{chiang2019cluster}, we evaluate
$$
(C,B) \in \{(15000,10), (1500,20), (200,1), (3000,100)\}.
$$
Since our hardware likely provides more available VRAM than was used in the original work, and to ensure a fair comparison with RNS, we additionally include a more memory-intensive ClusterGCN setting.

For GraphSAINT, following the configurations considered in the original paper~\cite{zeng2019graphsaint}, we evaluate
$$
(L_{\mathrm{w}}, S) \in \{(2,6000), (2,2000), (4,2000)\}.
$$
For the same reason, we also include an additional, more memory-intensive GraphSAINT setting.

For LADIES, we evaluate two configurations of the layer-wise sampling budget
vector $\mathbf{s}$ and batch size $B$:
$$
(\mathbf{s}, B)
\in
\left\{
\left([8192, 8192, 8192, 8192, 8192], 512\right),
\left([65536, 65536, 65536, 65536, 65536], 8192\right)
\right\}.
$$
The first configuration is chosen to remain close to the hyperparameter scale
used in the original LADIES paper~\cite{zou2019layer}. Since our experiments
consider substantially larger graphs, we also evaluate a larger sampling budget
and batch size to account for the increased graph scale.

For RNS, we evaluate
$$
m \in \{2,3,\dots,10\}.
$$

\subsubsection{Selected hyperparameters}
\label{subsec:selected_hparams}

The best configurations selected by the above tuning protocol are reported in Table~\ref{tab:hparams_sampling_compact}. In particular, the table should be read as the \emph{result of the full fairness pipeline described above}: for each dataset and method, we first searched over sampling hyperparameters, and for each candidate sampling configuration we separately optimized learning rate, dropout, and weight decay.

\begin{table*}[h]
\centering
\caption{Best hyperparameters selected for each dataset and sampling method. Sampling parameters are shown together with their corresponding tuned training hyperparameters. For LADIES, $\mathbf{s}=[a]^5$ denotes a five-layer sampling budget with $s_\ell=a$ for all layers.}
\label{tab:hparams_sampling_compact}
\begin{tabular}{llcccc}
\toprule
Dataset & Method & Sampling params & LR & Dropout & WD \\
\midrule
\multirow{6}{*}{\textsc{ogbn-arxiv}}
& Full       & --                          & $5\times10^{-4}$ & 0.5 & $5\times10^{-4}$ \\
& Neighbor   & $\mathbf{f}=[30,10,10]$     & $5\times10^{-4}$ & 0.2 & $10^{-3}$ \\
& ClusterGCN & $C=3000,\ B=100$            & $10^{-4}$        & 0.5 & $10^{-3}$ \\
& GraphSAINT & $L_{\mathrm{w}}=2,\ S=6000$ & $10^{-4}$        & 0.4 & $0$ \\
& LADIES     & $\mathbf{s}=[65536]^5,\ B=8192$ & $10^{-2}$    & 0.2 & $0$ \\
& RNS        & $m=2$                       & $5\times10^{-4}$ & 0.5 & $0$ \\
\midrule
\multirow{6}{*}{\textsc{ogbn-products}}
& Full       & --                          & $5\times10^{-4}$ & 0.3 & $5\times10^{-4}$ \\
& Neighbor   & $\mathbf{f}=[30,10,10,10]$  & $10^{-4}$        & 0.3 & $0$ \\
& ClusterGCN & $C=1500,\ B=20$             & $10^{-3}$        & 0.7 & $10^{-4}$ \\
& GraphSAINT & $L_{\mathrm{w}}=2,\ S=6000$ & $5\times10^{-4}$ & 0.6 & $0$ \\
& LADIES     & $\mathbf{s}=[8192]^5,\ B=512$ & $10^{-3}$     & 0.0 & $0$ \\
& RNS        & $m=10$                      & $5\times10^{-4}$ & 0.3 & $10^{-4}$ \\
\midrule
\multirow{6}{*}{\textsc{pokec}}
& Full       & --                          & $10^{-3}$        & 0.7 & $5\times10^{-4}$ \\
& Neighbor   & $\mathbf{f}=[30,10,10,10]$  & $10^{-4}$        & 0.3 & $0$ \\
& ClusterGCN & $C=1500,\ B=20$             & $10^{-4}$        & 0.3 & $5\times10^{-3}$ \\
& GraphSAINT & $L_{\mathrm{w}}=2,\ S=6000$ & $10^{-4}$        & 0.4 & $5\times10^{-4}$ \\
& LADIES     & $\mathbf{s}=[8192]^5,\ B=512$ & $5\times10^{-4}$ & 0.5 & $0$ \\
& RNS        & $m=2$                       & $10^{-3}$        & 0.3 & $10^{-4}$ \\
\bottomrule
\end{tabular}
\end{table*}

\subsection{Empirical evaluation of the implicit regularization with RNS}
\label{app:implicit-reg-impl-details}

We use the same experimental setup and hyperparameters as in the sampling benchmark. 
In particular, the data preprocessing, train/validation/test splits, model architectures, and evaluation protocol are kept unchanged for \textsc{ogbn-products}.

For RNS (no batch), we iterate over all mini-batches $B_k$ in the epoch, accumulate the corresponding losses gradients, and perform a single optimizer step after the full pass. This is equivalent to gradient accumulation over the entire epoch, and ensures that the update is computed from the same examples while removing the effect of stochastic mini-batch updates.

Unless otherwise stated, we do not use weight decay. Dropout is fixed to $0.5$ in all experiments. For SGD, we use a learning rate of $0.03$. For SGD with momentum, we use a learning rate of $0.03$ and momentum coefficient $0.9$. For Adam, we use a learning rate of $0.001$.

\subsection{GraphLand benchmark}
\label{app:graphland}

We follow the experimental protocol of GraphLand~\cite{bazhenov2024graphland} as closely as possible. In particular, we use the official Random Low and Random High splits, the same evaluation metrics, and the same training budget of 1000 epochs. For evaluation, we report accuracy on multi-class classification tasks and average precision on binary classification tasks. We also adopt the GraphSAGE architecture used in the GraphLand benchmark, which differs slightly from our default architecture: it consists of three graph convolutional layers, with two-layer MLPs inserted between successive graph convolutions. We use this architecture to ensure consistency with the original GraphLand benchmark.

For hyperparameter selection, we follow the same procedure as in the sampling benchmark. For each dataset and method, we run 50 trials using W\&B Bayesian search over the same hyperparameter space for dropout, weight decay, and learning rate. For RNS, we additionally tune the number of parts over $\{2,3,5,7,10\}$. Hyperparameter search results are given in Table \ref{tab:hyperparameters-graphland}.

These experiments are run on NVIDIA L40S GPUs with 48GB of memory. For memory-intensive settings, we use activation checkpointing: specifically, for full-graph training on \textsc{pokec-regions} and for RNS training on \textsc{web-fraud}. For full-graph training on \textsc{web-fraud} and \textsc{web-topics}, activation checkpointing alone is not sufficient; therefore, we use layer-parallel training across two L40S GPUs while keeping the training configuration otherwise unchanged.

\begin{table}[t]
\centering
\caption{GraphLand hyperparameters. For RNS, $K$ denotes the number of parts.}
\label{tab:hyperparameters-graphland}
\resizebox{\linewidth}{!}{
\begin{tabular}{llccccccc}
\toprule
Dataset & Reg. & \multicolumn{3}{c}{Full} & \multicolumn{4}{c}{RNS} \\
\cmidrule(lr){3-5}\cmidrule(lr){6-9}
& & LR & Drop. & WD & LR & Drop. & WD & $K$ \\
\midrule
\textsc{artnet-exp}    & RH & 0.0001 & 0.1 & 0.001  & 0.0001 & 0.1 & 0.001  & 2 \\
\textsc{artnet-exp}    & RL & 0.005  & 0.2 & 0.001  & 0.0001 & 0.3 & 0.001  & 2 \\
\textsc{city-reviews}  & RH & 0.0005 & 0.2 & 0.0001 & 0.0005 & 0.2 & 0.0005 & 3 \\
\textsc{city-reviews}  & RL & 0.0001 & 0.1 & 0.001  & 0.0001 & 0.1 & 0.001  & 2 \\
\textsc{hm-categories} & RH & 0.0005 & 0.2 & 0.0001 & 0.0001 & 0.2 & 0.0005 & 3 \\
\textsc{hm-categories} & RL & 0.0001 & 0.2 & 0.0005 & 0.0001 & 0.3 & 0.001  & 2 \\
\textsc{pokec-regions} & RH & 0.0005 & 0.0 & 0.0005 & 0.01   & 0.1 & 0.0    & 7 \\
\textsc{pokec-regions} & RL & 0.0005 & 0.0 & 0.0005 & 0.0005 & 0.0 & 0.0005 & 2 \\
\textsc{tolokers-2}    & RH & 0.005  & 0.0 & 0.001  & 0.0005 & 0.7 & 0.0    & 3 \\
\textsc{tolokers-2}    & RL & 0.0001 & 0.3 & 0.001  & 0.0005 & 0.7 & 0.001  & 3 \\
\textsc{web-fraud}     & RH & 0.0005 & 0.1 & 0.0    & 0.0001 & 0.1 & 0.0    & 2 \\
\textsc{web-fraud}     & RL & 0.0005 & 0.1 & 0.0    & 0.0005 & 0.2 & 0.0    & 2 \\
\textsc{web-topics}    & RH & 0.0005 & 0.1 & 0.0    & 0.0005 & 0.1 & 0.0    & 2 \\
\textsc{web-topics}    & RL & 0.0005 & 0.1 & 0.0    & 0.0005 & 0.2 & 0.0005 & 2 \\
\bottomrule
\end{tabular}
}
\end{table}

\subsection{Architecture Benchmark Parameters}
\label{subsec:architectures_details}

\begin{table}[t]
\centering
\caption{Architectural hyperparameters for GCN, GraphSAGE, and GAT.}
\label{tab:arch_hparams_gnn}
\small
\setlength{\tabcolsep}{5pt}
\begin{tabular}{llcccccc}
\toprule
Dataset & Model & Hidden & Layers & Output proj. & Residual & Norm \\
\midrule
\multirow{3}{*}{\textsc{ogbn-arxiv}}
& GCN       & 512 & 5 & yes & yes & BN \\
& GraphSAGE & 256 & 4 & yes & yes & BN \\
& GAT       & 256 & 5 & yes & yes & BN \\
\midrule
\multirow{3}{*}{\textsc{ogbn-products}}
& GCN       & 256 & 5 & no & yes & LN \\
& GraphSAGE & 256 & 5 & no & no  & LN \\
& GAT       & 256 & 5 & no & no  & LN \\
\midrule
\multirow{3}{*}{\textsc{pokec}}
& GCN       & 256 & 7 & yes & yes & BN \\
& GraphSAGE & 256 & 7 & yes & yes & BN \\
& GAT       & 256 & 7 & yes & yes & BN \\
\bottomrule
\end{tabular}
\end{table}

For GraphSAGE, GCN, and GAT, we use the same architectural hyperparameters as \citet{luo2024classic}. However, \citet{luo2024classic} does not report the SGFormer hyperparameters, so we instead adopt those from the original SGFormer paper \citet{wu2023sgformer}. Because \textsc{ogbn-products} is not included in \citet{wu2023sgformer}, we use for this dataset the hyperparameter configuration reported for Amazon2M, which is the closest large-scale benchmark considered in that work. This choice may partly explain discrepancies between our SGFormer results and those reported by \citet{luo2024classic}. Tables~\ref{tab:arch_hparams_gnn} and~\ref{tab:arch_hparams_sgformer} summarize the resulting architectural hyperparameters.

As in the sampling benchmark experiments, we tune training-related hyperparameters separately for each dataset-architecture pair using 50 Bayesian optimization runs with \textsc{Weights \& Biases} Sweeps. For RNS, we fix the number of parts to the best-performing configuration identified in the sampling benchmark for each dataset, namely 2 for \textsc{ogbn-arxiv}, 10 for \textsc{ogbn-products}, and 2 for \textsc{pokec}. The selected training hyperparameters are reported in Table~\ref{tab:tuned_hparams}.

All experiments are repeated over five random seeds: 42, 123, 456, 789, and 101112.

Unless otherwise specified, all models use \texttt{use\_input\_projection = false}. For GAT, the number of attention heads is 1 in all cases. For SGFormer, we use $\alpha = 0.5$, \texttt{use\_graph = true}, \texttt{use\_weight = true}, and \texttt{graph\_weight = 0.5} for all datasets.

On \textsc{ogbn-products} and \textsc{pokec}, GAT runs result in out-of-memory errors in our implementation. Although \citet{luo2024classic} report GAT results on these datasets, their public repository does not appear to contain run artifacts or scripts that clarify how these results were obtained. One possible explanation is that their experiments relied on RNS for example during evaluation, whereas our benchmark uses a unified end-to-end implementation and evaluation protocol across all methods. We therefore omit these GAT results rather than report numbers obtained under a different or unclear setting.

\begin{table}[t]
\centering
\caption{Architectural hyperparameters for SGFormer.}
\label{tab:arch_hparams_sgformer}
\small
\setlength{\tabcolsep}{5pt}
\begin{tabular}{lcccccc}
\toprule
Dataset & Hidden & Layers & Heads & Residual & Norm & GNN branch \\
\midrule
\textsc{ogbn-arxiv}    & 256 & 1 & 1 & yes & BN & GCN, hidden 256, layers 3 \\
\textsc{ogbn-products} & 256 & 1 & 2 & yes & BN & GCN, hidden 256, layers 3 \\
\textsc{pokec}         & 64  & 1 & 1 & yes & BN & GCN, hidden 64, layers 2 \\
\bottomrule
\end{tabular}
\end{table}

For SGFormer, the difference between our results and those of \citet{luo2024classic} may similarly stem from differences in the experimental setup. In particular, \citet{luo2024classic} do not report the SGFormer architectural hyperparameters, and their public repository does not appear to include the code or run artifacts needed to reproduce their SGFormer results on \textsc{ogbn-products}. Consequently, we report SGFormer results using the architecture specified above and the same unified training protocol used for the other models.

\begin{table}[t]
\centering
\caption{Tuned training hyperparameters for each dataset, architecture, and training setting. Each entry is reported as (dropout, learning rate, weight decay). N/A indicates settings for which runs did not yield valid results.}
\label{tab:tuned_hparams}
\small
\setlength{\tabcolsep}{5pt}
\begin{tabular}{llcc}
\toprule
Dataset & Architecture & Full & RNS \\
\midrule
\multirow{4}{*}{\textsc{ogbn-arxiv}}
& GCN       & $(0.5, 5 \times 10^{-4}, 0)$             & $(0.5, 5 \times 10^{-4}, 0)$ \\
& GraphSAGE & $(0.5, 5 \times 10^{-4}, 5 \times 10^{-4})$ & $(0.5, 5 \times 10^{-4}, 0)$ \\
& GAT       & $(0.3, 5 \times 10^{-4}, 0)$             & $(0.5, 5 \times 10^{-4}, 5 \times 10^{-4})$ \\
& SGFormer  & $(0.5, 5 \times 10^{-4}, 10^{-4})$       & $(0.7, 5 \times 10^{-4}, 0)$ \\
\midrule
\multirow{4}{*}{\textsc{ogbn-products}}
& GCN       & $(0.5, 3 \times 10^{-3}, 10^{-3})$       & $(0.3, 10^{-3}, 0)$ \\
& GraphSAGE & $(0.3, 5 \times 10^{-4}, 5 \times 10^{-4})$ & $(0.3, 5 \times 10^{-4}, 10^{-4})$ \\
& GAT       & N/A                                      & N/A \\
& SGFormer  & $(0.0, 10^{-2}, 0)$                      & $(0.7, 5 \times 10^{-3}, 0)$ \\
\midrule
\multirow{4}{*}{\textsc{pokec}}
& GCN       & $(0.2, 5 \times 10^{-4}, 10^{-5})$       & $(0.5, 10^{-3}, 0)$ \\
& GraphSAGE & $(0.7, 10^{-3}, 5 \times 10^{-4})$       & $(0.3, 10^{-3}, 10^{-4})$ \\
& GAT       & N/A                                      & N/A \\
& SGFormer  & $(0.0, 10^{-2}, 10^{-4})$                & $(0.7, 10^{-3}, 0)$ \\
\bottomrule
\end{tabular}
\end{table}

\section{Additional Results}

\subsection{RNS sampled graphs properties}

\begin{figure}[H]
  \centering
  \includegraphics[width=\textwidth]{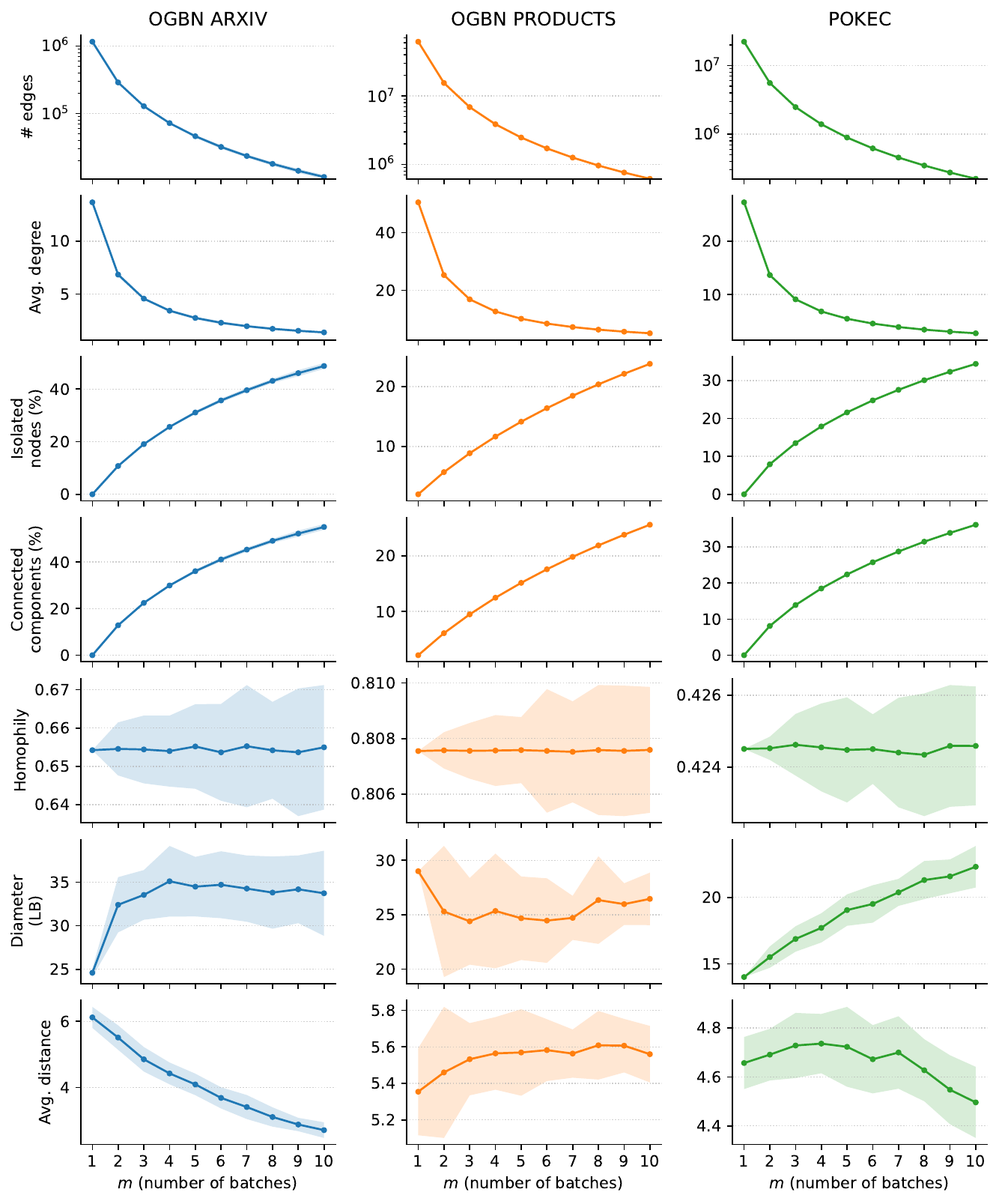}
  \caption{\textbf{Properties of sampled graphs as a function of the number of batches $m$.}
  Evolution of subsampled-graph metrics with respect to $m$ (number of batches per epoch). Shaded regions indicate $\pm$ one standard deviation over 5 epochs (full pass over all batches). The diameter lower bound corresponds to an estimate of the maximum distance between two nodes within the same connected component, computed using multi-sweep BFS. The average distance is an estimate of the expected shortest-path distance between two nodes within the same connected component, also computed using multi-sweep BFS.}
  \label{fig:sampling-properties}
\end{figure}

On Figure \ref{fig:sampling-properties}, we report connectivity metrics on sampled graphs as a function of the number of batches per epoch $m$.

\subsection{Speedup decomposition}
\label{app:rns_speedup}

\begin{table}[t]
\centering
\caption{\textbf{RNS speedup decomposition.}
Speedups over full-graph training for validation time-to-accuracy targets on a
single NVIDIA H100 GPU. Both methods reach every target in all five seeds.}
\label{tab:rns_speedup_decomposition}
\resizebox{\linewidth}{!}{
\begin{tabular}{llcccc}
\toprule
Dataset & Target val. & Epoch time & Epochs to target & Time to target & Time to target \\
        & accuracy    & speedup    & speedup           & w/ startup     & w/o startup \\
\midrule
\multirow{3}{*}{\textsc{ogbn-arxiv}}
  & 68.0\% & 1.07$\times$ & 2.02$\times$ & 1.67$\times$ & 2.13$\times$ \\
  & 69.0\% & 1.07$\times$ & 1.97$\times$ & 1.68$\times$ & 2.09$\times$ \\
  & 70.0\% & 1.07$\times$ & 1.92$\times$ & 1.69$\times$ & 2.06$\times$ \\
\midrule
\multirow{3}{*}{\textsc{ogbn-products}}
  & 88.0\% & 1.77$\times$ & 6.88$\times$ & 4.95$\times$ & 12.00$\times$ \\
  & 89.0\% & 1.77$\times$ & 6.87$\times$ & 5.59$\times$ & 11.99$\times$ \\
  & 90.0\% & 1.77$\times$ & 6.72$\times$ & 6.32$\times$ & 11.74$\times$ \\
\midrule
\multirow{3}{*}{\textsc{pokec}}
  & 78.0\% & 1.45$\times$ & 1.75$\times$ & 2.05$\times$ & 2.34$\times$ \\
  & 79.0\% & 1.45$\times$ & 1.51$\times$ & 1.88$\times$ & 2.03$\times$ \\
  & 80.0\% & 1.45$\times$ & 1.30$\times$ & 1.71$\times$ & 1.78$\times$ \\
\bottomrule
\end{tabular}
}
\end{table}

For each dataset, method, seed, and target validation accuracy $a^\star$, we
compute all quantities from validation logs. The per-epoch runtime is estimated
by the median slope between consecutive logs,
\[
\tau =
\operatorname{median}_j
\frac{\mathrm{runtime}_{j+1}-\mathrm{runtime}_j}
     {\mathrm{epoch}_{j+1}-\mathrm{epoch}_j}.
\]
Epochs-to-target $E(a^\star)$ are obtained by linear interpolation between the
last validation point below $a^\star$ and the first point above it. Applying the
same interpolation to runtime gives $T_{\mathrm{start}}(a^\star)$, and
$T_{\mathrm{nostart}}(a^\star)=T_{\mathrm{start}}(a^\star)-t_0$ removes startup
overhead, where $t_0$ is the first logged runtime.

We average run-level quantities over seeds and report speedups relative to
full-graph training:
\[
S_{\mathrm{epoch}} =
\frac{\mathbb{E}_s[\tau_{\mathrm{full},s}]}
     {\mathbb{E}_s[\tau_{\mathrm{RNS},s}]},
\quad
S_{\mathrm{epochs}} =
\frac{\mathbb{E}_s[E_{\mathrm{full},s}(a^\star)]}
     {\mathbb{E}_s[E_{\mathrm{RNS},s}(a^\star)]},
\quad
S_{\mathrm{time},\bullet} =
\frac{\mathbb{E}_s[T_{\bullet,\mathrm{full},s}(a^\star)]}
     {\mathbb{E}_s[T_{\bullet,\mathrm{RNS},s}(a^\star)]},
\]
for $\bullet\in\{\mathrm{start},\mathrm{nostart}\}$. Wall-clock speedups are
measured directly from runtime logs and therefore need not equal
$S_{\mathrm{epoch}}S_{\mathrm{epochs}}$, due to validation, logging,
data-loading, optimizer, and initialization overheads.

\subsection{Correlation between power-law preservation and batch gradient variance $R(w)$}
\label{app:alpha_R_correlation}

We run the following experiment on \textsc{ogbn-arxiv}, \textsc{ogbn-products}, and \textsc{pokec} at random
initialization, over 100 seeds and 50 batches per seed. For each seed and sampler, we
estimate the power-law exponent $\alpha_{\mathrm{Sampl}}$ by averaging Clauset MLE
estimates \cite{clauset2009power} across the 50 sampled subgraphs, and compute the gradient variance $R(w)$ from
the same batches. Each sampler yields one point in the
$\bigl(|\alpha_{\mathrm{Sampl}} - \alpha_{\mathrm{Full}}|,\, R(w)\bigr)$ plane. We used the optimal parameters found in the sampling benchmark for the estimations (See Appendix~\ref{sec:implementation_details}).

\begin{figure}[H]
    \centering
    \includegraphics[width=\linewidth]{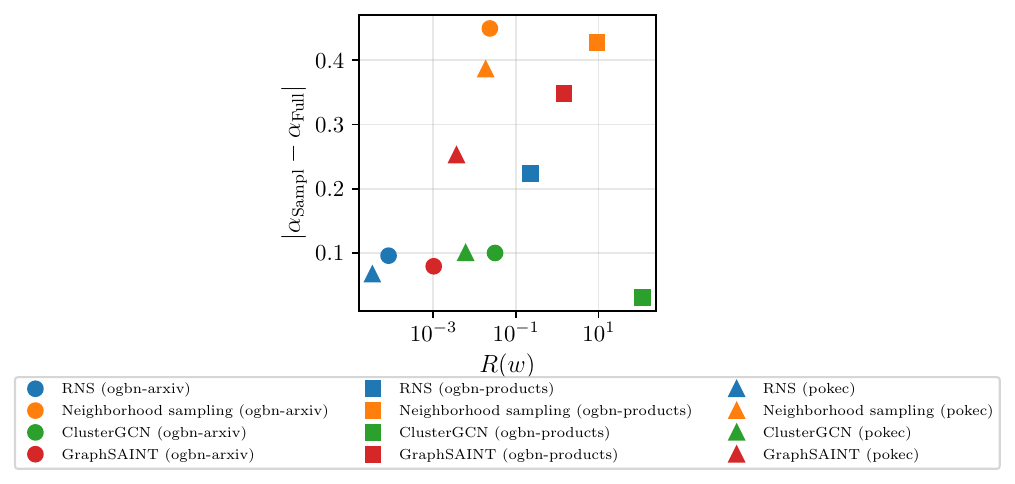}
    \caption{Power-law exponent error $|\alpha_{\mathrm{Sampl}} - \alpha_{\mathrm{Full}}|$
    vs.\ batch gradient variance $R(w)$ at random initialization, across samplers and
    datasets.}
    \label{fig:alpha_error_R}
\end{figure}

Figure~\ref{fig:alpha_error_R} shows a consistent positive relationship between the two
quantities: samplers that distort the degree distribution more also
induce higher gradient variance, with RNS in the bottom-left corner in every panel.
Intuitively, when every mini-batch faithfully reproduces the full-graph degree
distribution, batches are statistically homogeneous, so per-batch gradients are stable
estimators of $\nabla \bar{L}_{\mathrm{Sampl}}$ and $R(w)$ remains small. Samplers that
over- or under-represent high-degree nodes create batches with systematically different
gradient signals, inflating $R(w)$. While this correlation does not constitute a formal
proof, it bridges the graph-structural guarantee of Appendix~\ref{app:degree-thinning}
and the optimization term $R(w)$.

\subsection{Validation metrics through training}

\begin{figure}[H]
    \centering
    \includegraphics[width=0.65\linewidth]{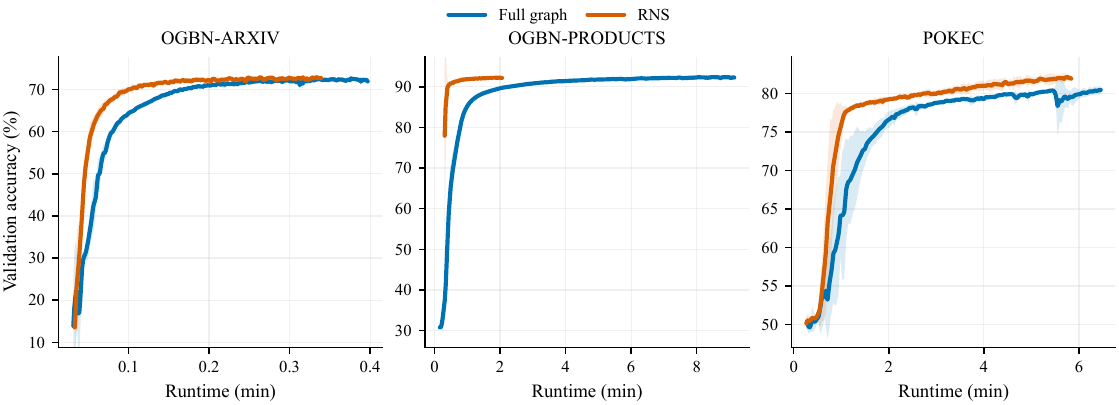}
    \caption{
    Validation accuracy over 200 epochs for full-graph training and RNS.
    }
    \label{fig:val_accuracy_training_dynamics}
\end{figure}

Figure~\ref{fig:val_accuracy_training_dynamics} reports validation accuracy over 200 training epochs for full-graph training and RNS. The curves show that the clear speed advantage of RNS in runtime. 

\subsection{Neighborhood sampling depth}
\label{sec:neighbor_sampling_depth}

We further analyze the effect of the neighbor sampling depth in
Table~\ref{tab:neighbor_fanout_ablation}. Although neighbor sampling is not the
main focus of our work, this ablation helps explain the fanout schedules used in
our benchmark and why we do not consider configurations deeper than four layers
(see Appendix~\ref{sec:implementation_details} for experimental details).
Increasing the depth can improve accuracy on larger graphs, especially
\textsc{pokec}, but this comes at a substantial computational cost. In our
implementation, PyG's neighbor sampler runs on the CPU, making deeper schedules
prohibitively slow for large-scale experiments.

\begin{table}[t]
\centering
\setlength{\tabcolsep}{5pt}
\caption{\textbf{Neighbor sampling fanout ablation.}
Test accuracy, time per epoch, and peak memory usage for different fanout
vectors. Results are reported as mean $\pm$ 95\% CI over 5 seeds. Increasing
the sampling depth generally improves accuracy on the larger graphs, but
substantially increases sampling cost and memory usage.}
\label{tab:neighbor_fanout_ablation}
\begin{tabular}{llcccc}
\toprule
Dataset & Fanout $\mathbf{f}$ & Accuracy (\%) & Time / Epoch (s) & GPU (GB) & RAM (GB) \\
\midrule
\multirow{3}{*}{\textsc{ogbn-arxiv}}
& $[30,10]$          & 71.02 $\pm$ 0.27 & 9.53  & 0.84  & 1.67 \\
& $[30,10,10]$       & \textbf{71.08 $\pm$ 0.47} & 20.43 & 1.61  & 1.67 \\
& $[30,10,10,10]$    & 70.95 $\pm$ 0.22 & 36.38 & 2.56  & 1.70 \\
\midrule
\multirow{3}{*}{\textsc{ogbn-products}}
& $[30,10]$          & 80.21 $\pm$ 0.35 & 58.72  & 1.85  & 10.07 \\
& $[30,10,10]$       & 80.44 $\pm$ 0.34 & 196.78 & 8.44  & 10.38 \\
& $[30,10,10,10]$    & \textbf{80.56 $\pm$ 0.28} & 638.15 & 24.59 & 10.51 \\
\midrule
\multirow{3}{*}{\textsc{pokec}}
& $[30,10]$          & 77.75 $\pm$ 0.09 & 40.56  & 1.89  & 4.67 \\
& $[30,10,10]$       & 80.07 $\pm$ 0.06 & 200.94 & 9.26  & 4.70 \\
& $[30,10,10,10]$    & \textbf{80.70 $\pm$ 0.13} & 729.76 & 29.92 & 4.85 \\
\bottomrule
\end{tabular}
\end{table}

%%%%%%%%%%%%%%%%%%%%%%%%%%%%%%%%%%%%%%%%%%%%%%%%%%%%%%%%%%%%

\newpage
\section*{NeurIPS Paper Checklist}

\begin{enumerate}

\item {\bf Claims}
    \item[] Question: Do the main claims made in the abstract and introduction accurately reflect the paper's contributions and scope?
    \item[] Answer: \answerYes{} % Replace by \answerYes{}, \answerNo{}, or \answerNA{}.
    \item[] Justification: The abstract and introduction clearly summarize the main claims and contributions made in the paper. 

\item {\bf Limitations}
    \item[] Question: Does the paper discuss the limitations of the work performed by the authors?
    \item[] Answer: \answerYes{} % Replace by \answerYes{}, \answerNo{}, or \answerNA{}.
    \item[] Justification: Section \ref{sec:limitations} points out two limitations of our work, namely the lack of theoretical characaterization of the implicit regularization term when optimizing with Adam, and the high influence of the number of batches per epoch $m$. 

\item {\bf Theory assumptions and proofs}
    \item[] Question: For each theoretical result, does the paper provide the full set of assumptions and a complete (and correct) proof?
    \item[] Answer: \answerYes{} % Replace by \answerYes{}, \answerNo{}, or \answerNA{}.
    \item[] Justification: Assumptions and full proofs are given in the Appendix (\ref{app:implicit_proof} and \ref{app:degree-thinning}).

    \item {\bf Experimental result reproducibility}
    \item[] Question: Does the paper fully disclose all the information needed to reproduce the main experimental results of the paper to the extent that it affects the main claims and/or conclusions of the paper (regardless of whether the code and data are provided or not)?
    \item[] Answer: \answerYes{} % Replace by \answerYes{}, \answerNo{}, or \answerNA{}.
    \item[] Justification: We believe that the detailed descriptions of all our experiments in Appendix \ref{app:experimental_details} are enough to reproduce each experimental results. We explicitly indicate which libraries and functions are used and when we reimplemented methods, we provide details on how we implemented the code.

\item {\bf Open access to data and code}
    \item[] Question: Does the paper provide open access to the data and code, with sufficient instructions to faithfully reproduce the main experimental results, as described in supplemental material?
    \item[] Answer: \answerYes{} % Replace by \answerYes{}, \answerNo{}, or \answerNA{}.
    \item[] Justification: Anonymous code is given with sufficient instructions to run the code.

\item {\bf Experimental setting/details}
    \item[] Question: Does the paper specify all the training and test details (e.g., data splits, hyperparameters, how they were chosen, type of optimizer) necessary to understand the results?
    \item[] Answer: \answerYes{} % Replace by \answerYes{}, \answerNo{}, or \answerNA{}.
    \item[] Justification: Experimental settings are mentioned in the corresponding paragraphs in the core paper. Then, all details are provided in Appendix \ref{app:experimental_details}.

\item {\bf Experiment statistical significance}
    \item[] Question: Does the paper report error bars suitably and correctly defined or other appropriate information about the statistical significance of the experiments?
    \item[] Answer: \answerYes{} % Replace by \answerYes{}, \answerNo{}, or \answerNA{}.
    \item[] Justification: We report mean performance together with 95\% confidence intervals for all experimental results. The intervals are computed across independent runs using different random seeds, capturing variability due to model initialization and stochastic training.

\item {\bf Experiments compute resources}
    \item[] Question: For each experiment, does the paper provide sufficient information on the computer resources (type of compute workers, memory, time of execution) needed to reproduce the experiments?
    \item[] Answer: \answerYes{} % Replace by \answerYes{}, \answerNo{}, or \answerNA{}.
    \item[] Justification: The paper clearly states what type of GPU was used for each experiment. While the description might not be complete, we believe that it is largely sufficient to reproduce the experiments.

\item {\bf Code of ethics}
    \item[] Question: Does the research conducted in the paper conform, in every respect, with the NeurIPS Code of Ethics \url{https://neurips.cc/public/EthicsGuidelines}?
    \item[] Answer: \answerYes{} % Replace by \answerYes{}, \answerNo{}, or \answerNA{}.
    \item[] Justification: We have reviewed the NeurIPS Code of Ethics and confirm that the research conducted in this paper conforms to it.

\item {\bf Broader impacts}
    \item[] Question: Does the paper discuss both potential positive societal impacts and negative societal impacts of the work performed?
    \item[] Answer: \answerNA{} % Replace by \answerYes{}, \answerNo{}, or \answerNA{}.
    \item[] Justification: The paper focuses on GNN training strategies and is primarily methodological. It does not introduce a new application, dataset involving people, deployment setting, or capability with a direct path to societal harm. As such, we do not identify direct positive or negative societal impacts.
    
\item {\bf Safeguards}
    \item[] Question: Does the paper describe safeguards that have been put in place for responsible release of data or models that have a high risk for misuse (e.g., pre-trained language models, image generators, or scraped datasets)?
    \item[] Answer: \answerNA{} % Replace by \answerYes{}, \answerNo{}, or \answerNA{}.
    \item[] Justification: The paper does not release data or models with a high risk of misuse.

\item {\bf Licenses for existing assets}
    \item[] Question: Are the creators or original owners of assets (e.g., code, data, models), used in the paper, properly credited and are the license and terms of use explicitly mentioned and properly respected?
    \item[] Answer: \answerYes{} % Replace by \answerYes{}, \answerNo{}, or \answerNA{}.
    \item[] Justification: We properly cite and credit the creators of all datasets and software libraries used in the paper. In particular, we use OGB datasets and GraphLand datasets, whose official repositories are released under the MIT license. Our implementation builds on PyTorch and PyTorch Geometric. PyTorch is released under a BSD-style license and PyTorch Geometric under the MIT license. We respect the corresponding license terms and terms of use.

\item {\bf New assets}
    \item[] Question: Are new assets introduced in the paper well documented and is the documentation provided alongside the assets?
    \item[] Answer: \answerNA{} % Replace by \answerYes{}, \answerNo{}, or \answerNA{}.
    \item[] Justification: No new asset is released.

\item {\bf Crowdsourcing and research with human subjects}
    \item[] Question: For crowdsourcing experiments and research with human subjects, does the paper include the full text of instructions given to participants and screenshots, if applicable, as well as details about compensation (if any)? 
    \item[] Answer: \answerNA{} % Replace by \answerYes{}, \answerNo{}, or \answerNA{}.
    \item[] Justification: The paper does not involve crowdsourcing nor research with human subjects.

\item {\bf Institutional review board (IRB) approvals or equivalent for research with human subjects}
    \item[] Question: Does the paper describe potential risks incurred by study participants, whether such risks were disclosed to the subjects, and whether Institutional Review Board (IRB) approvals (or an equivalent approval/review based on the requirements of your country or institution) were obtained?
    \item[] Answer: \answerNA{} % Replace by \answerYes{}, \answerNo{}, or \answerNA{}.
    \item[] Justification: The paper does not involve crowdsourcing nor research with human subjects.

\item {\bf Declaration of LLM usage}
    \item[] Question: Does the paper describe the usage of LLMs if it is an important, original, or non-standard component of the core methods in this research? Note that if the LLM is used only for writing, editing, or formatting purposes and does \emph{not} impact the core methodology, scientific rigor, or originality of the research, declaration is not required.
    %this research? 
    \item[] Answer: \answerNA{} % Replace by \answerYes{}, \answerNo{}, or \answerNA{}.
    \item[] Justification: The core method development in this research does not involve LLMs as any important, original, or non-standard components.

\end{enumerate}

\end{document}